\documentclass{article}

\PassOptionsToPackage{numbers,compress}{natbib}
\usepackage[main,final]{neurips_2026}

\usepackage[utf8]{inputenc}
\usepackage[T1]{fontenc}
\usepackage{hyperref}
\usepackage{url}
\usepackage{booktabs}
\usepackage{amsfonts}
\usepackage{amsmath}
\usepackage{amssymb}
\usepackage{amsthm}
\usepackage{nicefrac}
\usepackage{microtype}
\usepackage{xcolor}
\usepackage{algorithm}
\usepackage{algpseudocode}
\usepackage{graphicx}
\usepackage{subcaption}
\usepackage{xspace}
\usepackage{float}
\usepackage{listings}
\usepackage[most]{tcolorbox}
\tcbuselibrary{skins,listings,breakable,listingsutf8}
\usepackage{tablefootnote}
\usepackage{wrapfig}

\newtheorem{theorem}{Theorem}
\newtheorem{proposition}{Proposition}

\newcommand{\rebuttaledit}[1]{\textcolor{black}{#1}}

\title{DUET: Optimize Token-Budget Allocation for Reinforcement Learning with Verifiable Rewards}

\author{%
  Haoyu Hu$^{1,*}$ \quad
  Xuandong Zhao$^{2}$ \quad
  Xuhai ``Orson'' Xu$^{3,\dagger}$ \quad
  Nori Jacoby$^{1,\dagger}$ \\
  $^{1}$Cornell University \quad
  $^{2}$University of California, Berkeley \quad
  $^{3}$Columbia University \\
  $^{*}$\texttt{hh824@cornell.edu}\quad
  $^{\dagger}$Joint senior authors
}

\begin{document}

\maketitle

\begin{abstract}
Reinforcement learning with verifiable rewards (RLVR) generates hundreds of thousands of tokens per training step, with rollout generation dominating the computational cost. The overall token budget can be controlled along two main dimensions: (i) deciding which prompts to allocate rollouts to, and (ii) deciding how long each rollout should be. Prior work has generally controlled only one of these dimensions at a time. We show that jointly tuning both decisions under a shared compute budget improves both reasoning quality and wall-clock training time. We instantiate this view as \textbf{DU}al-controlled tok\textbf{E}n alloca\textbf{T}ion (DUET), a computationally efficient layer over GRPO that uses a lightweight pre-rollout surrogate of prompt informativeness to set how many rollouts each prompt receives, and a marker-gated abort rule with importance reweighting to set when to stop them.
On Qwen3-1.7B trained on MATH, DUET outperforms full-budget GRPO and the other three budget-aware baseline methods. DUET's advantage further generalizes to other benchmarks across math and coding, and is on par with the best baseline on the scientific Q\&A domain, while also achieving a $1.62\times$ wall-clock speedup.
More notably, using only 50\% of the token budget, DUET still outperforms all baseline methods at their full budget, achieving an even higher $2.51\times$ speedup over full-budget GRPO.
We verify the high performance of DUET on other backbone LLMs, including Qwen3-4B to 14B and Llama-3.2-3B-Instruct.
Notably, the gap between DUET and the strongest baseline \emph{widens} as the budget tightens, contrary to the usual pattern in which efficient methods trade off quality as compute decreases.
More broadly, these results suggest that DUET budget-aware control strategies are valuable not only for accelerating training, but also for improving the quality of the learning signal. \rebuttaledit{The code and data setup are publicly available at \href{https://github.com/Haoyu-Hu/DUET}{https://github.com/Haoyu-Hu/DUET}}

\end{abstract}
\vspace{-0.6cm}
\begin{figure}[H]
\centering
\includegraphics[width=0.95\linewidth]{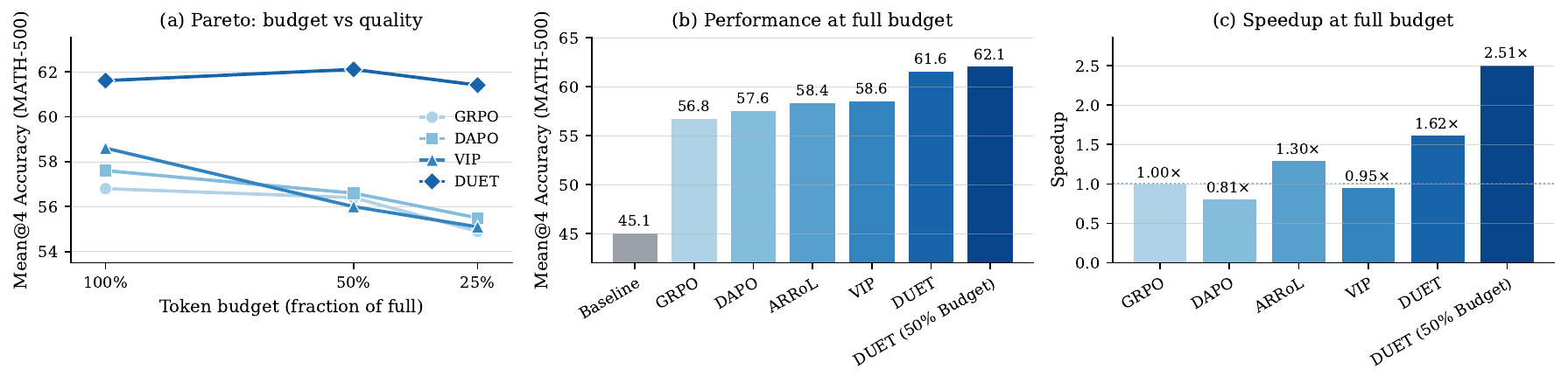}
\vspace{-0.3cm}
\caption{DUET dominates the budget--accuracy frontier on Qwen3-1.7B-base trained on MATH. (a)~Accuracy versus rollout budget for four methods at three budget points; DUET dominates at every point. (b)~Accuracy at full budget across baselines; DUET at half budget exceeds full-budget GRPO. (c)~Wall-clock speedup at full budget, normalized to GRPO; DUET at half budget runs $2.51\times$ faster than full-budget GRPO.}
\label{fig:money}
\end{figure}

\section{Introduction}\label{sec:intro}
\vspace{-0.25cm}
Reasoning-centric large language models (LLMs) such as DeepSeek-R1~\citep{deepseekai2025r1}, Light-R1~\citep{wen2025lightr1}, and Qwen3~\citep{qwen2025qwen3} have advanced state-of-the-art performance on mathematical and code-reasoning benchmarks~\citep{hendrycks2021math,cobbe2021gsm8k,chen2021humaneval}, and the post-training engine behind these results is reinforcement learning with verifiable rewards (RLVR)~\citep{shao2024deepseekmath}.  The recent rise of RLVR has been closely associated with GRPO~\citep{shao2024deepseekmath}, whose recipe is straightforward: at each training step, draw several candidate solutions per prompt, score them with a fast verifier or reward model, and update the policy using advantages computed relative to the rewards within that group. The recipe is also expensive: a single training step requires generating hundreds of thousands of tokens. Meanwhile, the verifier or reward model themselves are significantly cheaper~\citep{xu2026arrol,zheng2025greso}. Thus, as reasoning chains grow longer and models grow larger, the cost of an RLVR run is increasingly dominated by \emph{the total number of tokens} per training step.

A critical question follows: \emph{how do we reduce the token cost without losing model performance?} Recent work (Appendix~\ref{supp:related}) optimizes one of two natural degrees of freedom: either deciding which prompts should receive rollouts~\citep{yu2025dapo,zheng2025greso,nguyen2026vip,kong2025token} or deciding how long each rollout should continue~\citep{xu2026arrol}. Imposing a fixed length cap on every rollout is strictly suboptimal: aggressive truncation rewards the policy for committing to its first guess and silently eliminates the long chains of thought that underlie frontier reasoning performance~\citep{xiang2025justenough,yuan2025lazypenalty,yu2025dapo}. In this work we coordinate the prompt-level decision of how many rollouts to draw with the within-rollout decision of when to stop them, all under a single shared compute budget. This simple coordination delivers substantial gains over either dimension alone.

Classical statistics offers a clean solution to precisely this two-decision budget-allocation problem. Stratified Monte-Carlo integration~\citep{neyman1934representative,owen2013monte} and budgeted best-arm identification~\citep{evendar2006action,chen2000ocba} both treat ``how many samples'' and ``when to stop sampling'' as two faces of the same problem, governed by a single shared budget-pressure signal. We bring this perspective to RLVR with \textbf{\underline{DU}}al-controlled tok\textbf{\underline{E}}n alloca\textbf{\underline{T}}ion (\textbf{DUET}). At each training step, DUET estimates each prompt's informativeness from recent within-prompt gradient variance and allocates more rollouts to prompts that provide a stronger learning signal. During generation, DUET evaluates each rollout against two adaptive thresholds derived from the distribution of previously generated rollout lengths: rollouts that run too long without producing an answer are terminated as likely dead-end chains, while successful rollouts are stopped shortly after an answer appears. A single budget-pressure signal couples these two controls and enforces the compute budget at every step: when tokens are scarce, both rollout allocation and length gates tighten; when tokens are plentiful, both relax. DUET requires no auxiliary models trained online, adds less than one percent overhead to a standard GRPO step, and operates within the curated multi-pass training regime that dominates published RLVR practice~\citep{yu2025dapo,zeng2025simplerlzoo,wen2025lightr1,li2025limr,wang2025oneshotrlvr}.

Our proposal yields substantial performance gains (Figure~\ref{fig:money}). On Qwen3-1.7B-base~\citep{qwen2025qwen3} trained on MATH~\citep{hendrycks2021math}, DUET at half the rollout budget matches or beats every full-budget baseline (GRPO, DAPO, ARRoL, VIP) on all five reasoning benchmarks (MATH-500, GSM8K, AIME-2024, HumanEval, GPQA-Diamond) at $2.51\times$ wall-clock speedup over full-budget GRPO. At full budget, DUET surpasses these three budget-aware baselines (DAPO, ARRoL, VIP) on the same benchmarks at $1.62\times$ speedup. These gains are largely maintained from 4B to 14B scale and also transfer to a Llama-3.2-3B-Instruct~\citep{grattafiori2024llama3}. \rebuttaledit{We further verified the robustness and efficiency of DUET via comprehensive ablation studies including asynchronous RL methods and different rollout allocation heuristics. These findings show that these gains are not merely a flat speedup.}

 \emph{DUET learns which prompts merit compute without supervision.} The per-prompt rollout count begins uniformly at $4$ and expands to a range as wide as $[1,32]$ as the controller’s online variance estimate accumulates, with accuracy increasing fastest during this fan-out phase. \emph{DUET removes two forms of token waste under a single shared budget.} A marker-gated abort terminates rollouts that exceed the policy’s natural stopping length without producing a parsable answer, and trims successful rollouts shortly after the answer appears; both decisions are coupled by the same budget-pressure signal that enforces the budget. \emph{DUET’s advantage grows as the budget tightens.} Existing efficient-RLVR methods tend to trade quality for compute savings, so their gains shrink at lower budgets; DUET exhibits the opposite pattern, with its largest margin over the next-best baseline at quarter budget.
\vspace{-0.25cm}

\section{Problem Setup}\label{sec:setup}
\vspace{-0.1cm}
At each step $t$, an RLVR trainer draws a batch $\mathcal{B}_t$ of $M$ prompts, generates $n_q$ rollouts per prompt $q$ from the current policy $\pi_\theta$, scores them with a verifier, and updates the parameters $\theta$ to maximise $J(\theta) = \mathbb{E}_{q\sim\mu,\,y\sim\pi_\theta(\cdot\mid q)}[R(q,y)]$. The dominant computational cost of this step is the rollout generation: with expected rollout length $\bar L_q$, the per-step token count is $\sum_q n_q\,\bar L_q$, which we cap at a budget $B$. Under independent rollouts within a prompt, the trace variance of the standard policy-gradient estimator has the textbook form
\begin{equation}
V(n) \;=\; \sum_q \sigma_q^2/n_q,
\label{eq:setup-variance}
\end{equation}
where $\sigma_q^2$ is the within-prompt variance of the per-rollout gradient contribution $Z_{q,i} := A_{q,i}\,\nabla_\theta\log\pi_\theta(y_{q,i}\mid q)$, where $A_{q,i}$ is the advantage of rollout $i$ for prompt $q$.

The control problem is then simple: pick a per-prompt rollout count $n_q$ and a per-rollout stopping rule under the token budget. The two choices must keep $V(n)$ small and the gradient unbiased for the standard policy gradient. We restrict to the multi-pass curated-corpus regime ($E\ge 2$ epochs over a fixed prompt pool), which covers most published RLVR practice~\citep{yu2025dapo,zeng2025simplerlzoo,wen2025lightr1,hu2025orz,li2025limr,wang2025oneshotrlvr}. Statements below are written for an action-independent-baseline surrogate $g^{\text{AIB}}(\theta_t)$; the residual coupling under the GRPO group-normalised advantages used in the implementation is reported empirically and analysed in Appendix~\ref{supp:t2-proof}.

\section{DUal-controlled tokEn allocaTion (DUET)}\label{sec:method}

\begin{figure}[t]
\centering
\includegraphics[width=0.8\linewidth]{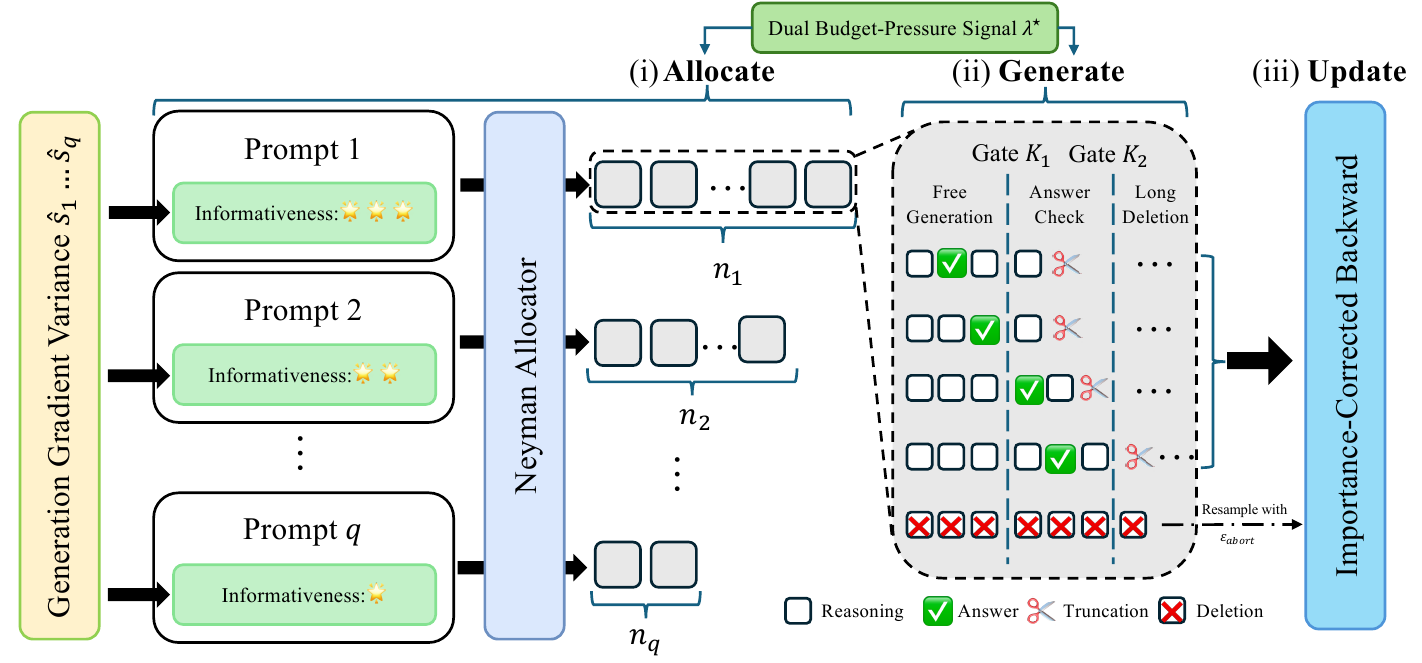}
\caption{One DUET training step. \emph{Allocate}: each prompt receives an informativeness score $\hat s_q$ from previous training contributions, and the cost-weighted Neyman rule sets the per-prompt rollout count $n_q$. \emph{Generate}: a marker-gated rule terminates a rollout shortly after an answer marker fires and aborts a marker-less rollout past a quantile threshold. \emph{Update}: kept rollouts feed an importance-corrected, gradient-masked backward step. The dual variable $\lambda^\star$ is the budget-pressure signal that closes the shared token budget across all three phases.}
\label{fig:method}
\end{figure}

DUET is a thin three-phase layer over the standard GRPO loop. \emph{Allocate} (Figure~\ref{fig:method}(i)) decides how many rollouts each prompt receives. \emph{Generate} (Figure~\ref{fig:method}(ii)) decides when each rollout stops. \emph{Update} (Figure~\ref{fig:method}(iii)) reweights the kept rollouts so the gradient stays unbiased. A single budget-pressure signal $\lambda^\star$ ties the three phases together and constrains the token budget at every step. The controller's overhead (a bisection over $\lambda$, a streaming marker check, and an importance reweighting on kept rollouts) adds less than one percent to per-step wall-clock at the model scales we run. The full pseudocode is Algorithm~\ref{alg:duet} in Appendix~\ref{supp:algorithm}.

\subsection{ Allocate: Cost-Weighted Neyman with an Online Surrogate}\label{sec:method-alloc}

This step concerns how many rollouts each prompt should receive under the per-step token budget $B$. Stratified Monte-Carlo integration~\citep{owen2013monte} provides exactly the recipe with variance minimization: sample more from prompts whose rollouts disagree more. The classical Neyman optimum is $n_q\propto\sigma_q$ when every rollout costs the same~\citep{neyman1934representative}. Rollouts in RLVR have prompt-dependent expected length $\bar L_q$, so the answer acquires a square-root cost factor.

\begin{theorem}[Cost-weighted Neyman allocation; proof in Appendix~\ref{supp:t1-proof}]\label{thm:t1}
For independent rollouts, among positive allocations $n$ satisfying $\sum_q n_q\,\bar L_q = B$, the variance $V(n) = \sum_q\sigma_q^2/n_q$ is uniquely minimised by
\begin{equation}
  n_q^\star \;=\; \frac{\sigma_q}{\sqrt{\lambda^\star\,\bar L_q}},
  \label{eq:t1-allocation}
\end{equation}
where $\lambda^\star > 0$ is the unique Lagrange multiplier closing the budget, and the minimum is $V^\star = \lambda^\star B$.
\end{theorem}

Theorem~\ref{thm:t1} is the basis for the following steps.  The numerator $\sigma_q$ tells the controller to spend more rollouts where the within-prompt rollouts disagree the most, since those are the prompts whose gradient signal is updating the policy. Its denominator $\sqrt{\bar L_q}$ favours cheap prompts, since a long rollout costs more tokens. The Lagrange multiplier $\lambda^\star$ is the shared budget-pressure signal: when tokens are scarce $\lambda^\star$ is large and every $n_q^\star$ shrinks; when tokens are plentiful $\lambda^\star$ is small and every $n_q^\star$ grows. Once the abort policy of Section~\ref{sec:method-abort} fixes the realized lengths $\bar L_q(\theta)$, a unique $\lambda^\star(\theta)$ constrains the budget under the abort policy (Proposition~\ref{prop:t3} in Appendix~\ref{supp:t3-proof}). The same scalar reappears in the abort gate of phase~(ii), which is the mechanism that ties the two phases under one shared budget.

The variance signal $\sigma_q$ is unobservable at allocation time. We use a per-prompt running estimate: at the end of every step, every prompt $q$ with $n_q\ge 2$ kept-not-aborted rollouts contributes
\begin{equation}
\hat\sigma^{\text{obs}}_{q,t} \;:=\; \operatorname{std}_i\!\Bigl(A_{q,i}\sum_\ell\log\pi_{\theta_t}(y_{q,i,\ell}\mid q, y_{q,i,<\ell})\Bigr),
\label{eq:method-sigma-obs}
\end{equation}
which updates a running mean $\hat\sigma^{\text{obs}}_q$. The plug-in surrogate is $\hat s_q := \max\{s_{\text{floor}},\hat\sigma^{\text{obs}}_q\}$ with a small cold-start floor for prompts not yet observed; the cost factor $\hat L_q$ is the per-prompt running mean over kept-rollout lengths. Substituting $\hat s_q$ into Theorem~\ref{thm:t1} keeps the allocation rule unchanged. The resulting plug-in cost is bounded above by a single divergence between surrogate and truth that decays as $1/B$ (Theorem~\ref{thm:t4} in Appendix~\ref{supp:t4-proof}). The empirical difference between $\hat s_q$ and the within-step variance estimate is therefore the natural diagnostic of allocation quality. The realised allocation is
\begin{equation}
n_q \;=\; \max\!\left( n_{\min},\;\operatorname{round}\!\left(\frac{\hat s_q}{\sqrt{\lambda^\star\,\hat L_q}}\right)\right),
\label{eq:method-alloc}
\end{equation}
where $\lambda^\star$ solves $\sum_{k\in\mathcal{B}_t} n_k(\lambda)\,\hat L_k = B$ via a few bisection iterations warm-started from $\lambda^\star_{t-1}$, and $n_{\min}$ enforces a per-prompt minimum.

\subsection{Generate: Marker-Gated Abort with $\varepsilon$-Keep}\label{sec:method-abort}

This step concerns controlling when each rollout should stop. A fixed-length cap is the simplest answer and the worst one: it penalizes a coherent partial chain-of-thought identically to a wrong answer, and silently kills the long traces that drive frontier reasoning~\citep{xiang2025justenough,yuan2025lazypenalty,yu2025dapo}. \rebuttaledit{Instead, we use an answer-specific stopping criterion. The marker $m(y^{1:t})\in\{0,1\}$ detects a parsable answer in the prefix, which reuses the machine veriable format defined by RLVR. For example, it could be a closed numerical box for math, a closing code fence for code, or an answer span for short-form QA.}

Two thresholds set the schedule. A polling threshold $K_1$ is the earliest token at which the marker check turns on; before $K_1$ the marker is rarely needed and inspection wastes cycles. An abort threshold $K_2$ is a high quantile (e.g. 80 \%) of the policy's natural-stopping length with our assumption that a rollout reaching $K_2$ without a marker is unlikely to land an answer. Both thresholds adapt online from a sliding window of recent kept rollouts. With $\tau_q^{\text{marker}} := \inf\{t \ge K_1 : m(y_{q,i,1:t})=1\}$ and a small grace window $G$ for a clean tail,
\begin{equation}
  \tau_q^{\text{abort}} \;:=\; \min\bigl\{\tau_q^{\text{marker}} + G,\; K_2 + G\bigr\}.
  \label{eq:abort-time}
\end{equation}

A rollout that emits a marker by $\tau_q^{\text{abort}}$ is kept and contributes its gradient at propensity $1$; this is the answer-fired path between $K_1$ and $K_2$ in phase~(ii). A marker-less rollout cannot simply be dropped, since dropping silently removes the slowest, least-confident prompts from the gradient and biases learning toward the prompts the policy already solves. We apply a minimal importance correction instead: with probability $\varepsilon_{\text{abort}}\in(0,1)$ the marker-less rollout is continued to natural EOS (end of sentence, subject to a large safeguard length limit) and its contribution is reweighted by $1/\varepsilon_{\text{abort}}$, and otherwise the rollout is aborted and masked out. Let $I_{q,i}^{\text{abort}}\in\{0,1\}$ be the abort-and-mask indicator and $p_{q,i}^{\text{abort}}\in\{1,\varepsilon_{\text{abort}}\}$ the kept-rollout propensity. The per-rollout estimator $\hat h_{q,i} := (1 - I_{q,i}^{\text{abort}})\,Z_{q,i}/p_{q,i}^{\text{abort}}$ is exactly unbiased for the natural-rollout contribution under an action-independent baseline. Its trace second moment lifts by at most $(1-p_q^{\text{marker}})(1-\varepsilon_{\text{abort}})/\varepsilon_{\text{abort}}$ times the marker-less stratum's second moment $\sigma^2_{\text{m-less},q}$ (Theorem~\ref{thm:t2} in Appendix~\ref{supp:t2-proof}). The surcharge is the price the controller pays to keep the slow prompts in the gradient. Under realisable optima it goes to zero asymptotically (Appendix~\ref{supp:self-extinguishing}); under KL regularisation it stays small but non-zero, and in our runs it never exceeds $1.5\times$ the natural-rollout second moment (Figure~\ref{fig:theory-empirics}(c)).

\subsection{Update: Importance-Corrected, Gradient-Masked Backward}\label{sec:method-update}

The last step runs the gradient update on the kept rollouts. Writing $w_{q,i} := (1 - I_{q,i}^{\text{abort}})/(s_q^{\text{pre}}\,p_{q,i}^{\text{abort}})$ with $s_q^{\text{pre}} := \operatorname{clip}(n_q/\bar n_t, \varepsilon_{\text{pre}}, 1)$ a per-prompt stratification factor and $m_{q,i,\ell}$ the post-abort response mask, where $\bar n_t = M^{-1}\sum_q n_q$ is the batch-mean rollout count at step $t$, and $\varepsilon_{\text{pre}}$ is a small lower clip floor,
\begingroup\color{black}\small
\begin{equation}
\mathcal{L}_t(\theta)=-\frac{1}{N_t}\sum_{q,i,\ell}w_{q,i}m_{q,i,\ell}\min\!\left\{r_{q,i,\ell}(\theta)A_{q,i},\operatorname{clip}\!\left(r_{q,i,\ell}(\theta),1-\epsilon_{\rm clip},1+\epsilon_{\rm clip}\right)A_{q,i}\right\}.
\label{eq:duet-loss}
\end{equation}
\endgroup
\rebuttaledit{where
\begin{equation}
    r_{q,i,\ell}(\theta)=\frac{\pi_\theta(y_{q,i,\ell}\mid q,y_{q,i,<\ell})}{\pi_{\theta_{\rm old}}(y_{q,i,\ell}\mid q,y_{q,i,<\ell})}, N_t=\sum_{q,i,\ell}m_{q,i,\ell}, \epsilon_{\rm clip}=0.2
\end{equation}
The response mask removes tokens after an abort. Group-normalised advantages and token-mean aggregation match the standard GRPO loss aggregator; DAPO retains its own $0.20/0.28$ clipping setting. The rollout count from \eqref{eq:method-alloc} and the abort rule from \eqref{eq:abort-time} remain coupled through the same dual $\lambda^\star$ that closes the budget.}

\section{Results}\label{sec:results}

\subsection{Experimental Setup}\label{sec:experiments}

\paragraph{Models and Datasets.}
Training uses Qwen3-1.7B-base, Qwen3-4B-base~\citep{qwen2025qwen3}, and Llama-3.2-3B-Instruct~\citep{grattafiori2024llama3} as a cross-family check, all with full-parameter fine-tuning on the Hendrycks MATH~\citep{hendrycks2021math} 7.5k corpus. Evaluation covers MATH-500~\citep{hendrycks2021math}, GSM8K~\citep{cobbe2021gsm8k}, and AIME-2024 for math; HumanEval~\citep{chen2021humaneval} for code generation; and GPQA-Diamond~\citep{rein2023gpqa} as a cross-domain reasoning probe. Full configurations appear in Appendix~\ref{supp:experimental-setup}. Due to the limitation of computing resources, we only test smaller models in this study.

\paragraph{Baselines and Protocol.}
We compare DUET against the non-budgeted reference (GRPO~\citep{shao2024deepseekmath}) and three efficient-RLVR baselines: DAPO~\citep{yu2025dapo} (dynamic-sampling), ARRoL~\citep{xu2026arrol} (within-rollout-pruning), and VIP~\citep{nguyen2026vip} (budget-adaptive), each at full, half and quarter rollout budget. \rebuttaledit{We additionally evaluate APRIL~\citep{zhou2025april}, an asynchronous partial-rollout method that reduces idle time from long-tail responses, and GRESO~\citep{zheng2025greso}, a reward-history prompt-skipping method, in Appendix~\ref{supp:robustness}.}  We used 8 as the full rollout budget per prompt, which is the most common choice for RL training~\citep{zheng2025greso,xu2026arrol}, and set the hard limit of the budget by capping the number of rollouts for methods without budget settings (GRPO, DAPO, ARRoL). All cells run on the same software and hardware. Specifically, we use a server with 8$\times$H100 80GB GPUs and verl 0.4.1~\citep{sheng2024hybridflow} with vLLM 0.9.2~\citep{kwon2023pagedattention}. We set top-$p$ to $0.95$, the max prompt length to $512$ tokens, and the max generation length to $3{,}072$ tokens. We train the model for $232$ steps ($\sim$4 epochs) on the MATH dataset, with each step using $128$ prompts. In the main content, we evaluate every $30$ training steps, generate $4$ rollouts per prompt at temperature $0.9$, and report the best mean@4 across checkpoints.
\rebuttaledit{Multi-seed trajectories, held-out checks, and scale-up runs are reported as robustness ablations in Appendix~\ref{supp:robustness}.}

\paragraph{Validating the rollout-budget knob.}
GRPO, DAPO, and VIP all expose a per-prompt rollout count as their primary budget control. Halving this knob halves the per-step generated-token cost on our setup (Figure~\ref{fig:token-decomposition} in Appendix~\ref{supp:quarter-arrol}), so ``rollout budget $b$'' in this paper is a faithful proxy for ``token budget $bB$''. ARRoL is the exception: its $\kappa$ knob is a soft post-generation drop probability, not a hard token cap, so we compare ARRoL only at its default operating point and audit its budget semantics in Appendix~\ref{supp:quarter-arrol}.

\begin{table}[t]
\caption{Accuracy (mean@4, \%) and GRPO-normalized speedup on Qwen3-1.7B-base, Qwen3-4B-base, and Llama-3.2-3B-Instruct trained on MATH. \textbf{Bold} = best per column within each model section; \underline{underline} = second-best (ties at the rounded value are all underlined). \rebuttaledit{Multi-seed and scale-up robustness checks are in Appendix~\ref{supp:robustness}, and full budgeted results are in Table~\ref{tab:quarter}.}}
\label{tab:main}
\centering
\small
\begin{tabular}{lcccccc}
\toprule
Method & MATH-500 & GSM8K & AIME-24 & HumanEval & GPQA-D & Speedup \\
\midrule
\multicolumn{7}{l}{\textit{Qwen3-1.7B-base}} \\
\midrule
Baseline             & 45.1 & 60.4 & 5.8 & 34.5 & 22.2 & --  \\
+GRPO                & 56.8 & 78.0 & 6.7 & 47.6 & 28.8 & 1$\times$ \\
+DAPO                & 57.6 & 77.4 & 7.5 & 46.8 & \textbf{29.1} & 0.81$\times$ \\
+ARRoL               & 58.4 & 79.6 & 8.3 & 47.7 & 26.7 & 1.30$\times$ \\
+VIP                 & 58.6 & 78.1 & 5.8 & 48.9 & 27.8 & 0.95$\times$ \\
+DUET (50\% Budget)  & \textbf{62.1} & \underline{84.0} & \underline{10.8} & \underline{51.7} & \textbf{29.1} & \textbf{2.51$\times$} \\
+DUET                & \underline{61.6} & \textbf{84.7} & \textbf{12.5} & \textbf{52.7} & \underline{29.0} & \underline{1.62$\times$} \\
\midrule
\multicolumn{7}{l}{\textit{Qwen3-4B-base}} \\
\midrule
Baseline             & 54.3 & 58.5 & 10.0 & 62.8 & 28.8 & -- \\
+GRPO                & 74.5 & 90.1 & 13.3 & 73.6 & \textbf{38.5} & 1$\times$ \\
+DAPO                & 73.2 & 89.9 & \textbf{18.3} & \underline{75.2} & 37.4 & 0.63$\times$ \\
+ARRoL               & 73.9 & 90.3 & 11.7 & 73.6 & 34.5 & 1.19$\times$ \\
+VIP                 & 74.4 & 90.1 & \underline{16.7} & \underline{75.2} & 36.8 & 0.75$\times$ \\
+DUET (50\% Budget)  & \underline{76.7} & \underline{92.3} & 13.3 & \textbf{77.0} & 38.1 & \textbf{2.38$\times$} \\
+DUET                & \textbf{78.2} & \textbf{93.9} & 14.2 & \textbf{77.0} & \underline{38.3} & \underline{1.43$\times$} \\
\midrule
\multicolumn{7}{l}{\textit{Llama-3.2-3B-Instruct}} \\
\midrule
Baseline             & 37.1 & 60.3 & 3.3 & 50.5 & 20.5 & -- \\
+GRPO                & 44.4 & 75.7 & \underline{7.5} & 52.8 & \textbf{25.4} & 1$\times$ \\
+DAPO                & 44.1 & 75.5 & \underline{7.5} & 52.7 & 24.5 & 0.82$\times$ \\
+ARRoL               & \underline{45.3} & 74.8 & 6.7 & 52.8 & \underline{25.3} & 1.13$\times$ \\
+VIP                 & 43.6 & 75.2 & \textbf{9.2} & \textbf{53.0} & 24.9 & 0.95$\times$ \\
+DUET (50\% Budget)  & 44.5 & 72.8 & 6.7 & \underline{52.9} & 23.2 & \textbf{2.04$\times$} \\
+DUET                & \textbf{45.6} & \textbf{75.9} & \underline{7.5} & 52.1 & \underline{25.3} & \underline{1.26$\times$} \\
\bottomrule
\end{tabular}
\end{table}
\vspace{-0.2cm}

\subsection{Main Results}\label{sec:main-results}

\paragraph{DUET matches full-budget GRPO quality at half the rollouts and half the wall-clock, and the headline transfers across model scale and model family.} Table~\ref{tab:main} reports mean@4 accuracy on the five evaluation benchmarks at two Qwen3 scales and on the Llama-3.2-3B-Instruct cross-family check. At half rollout budget, DUET matches or exceeds full-budget GRPO on four of five benchmarks at both Qwen3 scales while running $2.51\times$ (1.7B) and $2.38\times$ (4B) faster, so a practitioner reaches the same model in less than half the wall-clock. At full budget DUET leads four of five benchmarks at 1.7B and three of five at 4B, with a $1.62\times$ and $1.43\times$ speedup over same-rollout-count GRPO. The single column where vanilla GRPO leads on Qwen3 is GPQA-Diamond, an out-of-distribution probe for math-trained policies, and the gap is at most $0.5\%$. On Llama-3.2-3B-Instruct the RLHF prior compresses headroom across all methods, but full-budget DUET still leads on MATH-500 and GSM8K, and half-budget DUET runs $2.04\times$ faster than GRPO with at most a $1\%$ gap on every benchmark. HumanEval lifts by at most $2.5\%$ across every method on this family, a property of the strong RLHF prior rather than the training rule.
\vspace{-0.25cm}

\clearpage
\subsection{Wall-Clock Efficiency}\label{sec:efficiency}
\begin{wrapfigure}{r}{0.4\linewidth}
\centering
\vspace{-0.25cm}
\includegraphics[width=\linewidth]{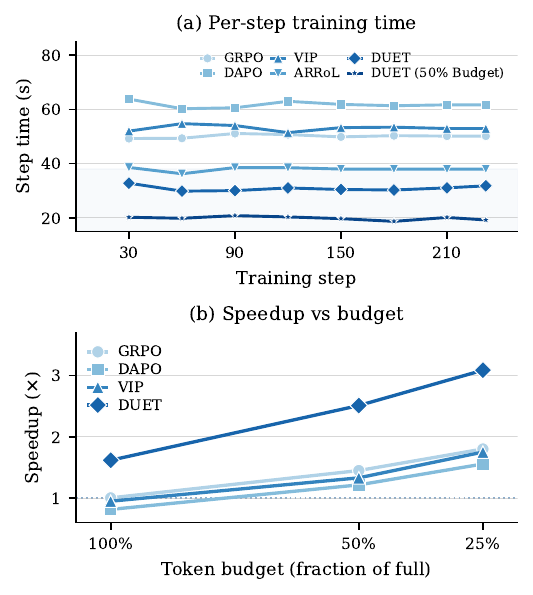}
\caption{\rebuttaledit{Wall-clock efficiency on Qwen3-1.7B / MATH. \textbf{(a)} Per-step training time. \textbf{(b)} Speedup over full-budget GRPO across rollout budgets.}}
\label{fig:step-time}
\vspace{0pt}
\end{wrapfigure}

\paragraph{DUET shifts the entire speed-quality Pareto outward, not just one operating point.} Figure~\ref{fig:step-time}(a) shows that DUET at full budget runs faster than full-budget GRPO from the first validation checkpoint onward, and DUET at half budget sits below every baseline throughout training, stabilising near a $2.5\times$ ratio once the online length-quantile estimator settles. Figure~\ref{fig:step-time}(b) sweeps the rollout budget at three points and reports speedup against full-budget GRPO: DUET dominates at every point, reaches $3.4\times$ at quarter budget, and pulls away from a much shallower GRPO/DAPO/VIP frontier. The separation is structural rather than incremental. ARRoL's pruner saves only the small slice of generated tokens past its mid-decode inspection point. DAPO's post-rollout filter pays the full generation cost on the rollouts it eventually drops. DUET's marker-gated abort acts before generation reaches the model's natural maximum length, and the savings compound across the heterogeneous allocation that the Neyman optimum prescribes. The Pareto curve does not just slide down. The \emph{slope} changes, so the practitioner's case for DUET strengthens at every tighter operating point.
\subsection{Ablations}\label{sec:ablation}

We first isolate two questions on DUET ablations with Qwen3-1.7B: do the allocator and the abort each pull their weight, and how sensitive is the performance to hyperparameters?\begin{table}[H]
\caption{Ablation of the cost-weighted Neyman allocator and the marker-gated abort on Qwen3-1.7B / MATH at $50\%$ rollout budget, $\varepsilon_{abort}$=0.05.}
\label{tab:ablation}
\centering
\small
\begin{tabular}{lcccccc}
\toprule
Configuration            & MATH-500 & GSM8K & AIME-24 & HumanEval & GPQA-D & Speedup \\
\midrule
Baseline       & 45.1 & 60.4 & 5.8 & 34.5 & 22.2 & --  \\
GRPO & 56.4 & 75.8 & 6.7 & 46.9 & 27.4 & 1$\times$ \\
+ allocator only         & \underline{61.7} & \textbf{84.1} & \textbf{12.5} & \underline{51.0} & 28.3 & 1.30$\times$ \\
+ abort only             & 61.3 & 82.1 & \textbf{12.5} & 49.4 & \underline{28.6} & \textbf{1.93$\times$} \\
+ both (DUET)            & \textbf{62.1} & \underline{84.0} & \underline{10.8} & \textbf{51.7} & \textbf{29.1} & \underline{1.73$\times$} \\
\bottomrule
\end{tabular}
\end{table}

\paragraph{The allocator and the abort specialise; combining them is Pareto-better than either alone.} Table~\ref{tab:ablation} disables one ingredient at a time. The allocator alone (uniform abort off) delivers $+5.3\%$ on MATH-500 over the rollout-matched anchor at $1.30\times$ speedup, and is the per-benchmark winner on GSM8K and AIME-24; the abort alone (uniform $n_q$, abort on) delivers $+4.9\%$ at $1.93\times$ speedup. The two ingredients are not interchangeable: the allocator concentrates compute on prompts whose rollouts move the gradient, while the abort cuts the marker-less tail of every rollout. Combining them into the full controller produces $+5.7\%$ on MATH-500 (a Pareto improvement on quality) at $1.73\times$ speedup, and the highest HumanEval and GPQA-Diamond scores of the four cells. The allocator is the quality lever, the abort is the wall-clock lever, and the joint controller takes both at once. 

\rebuttaledit{The broader evidence points in the same direction. Three-seed final and best-checkpoint trajectories prove that DUET converge not only faster but also higher. The performance maintains on two held-out benchmarks, Minerva-Math~\citep{lewkowycz2022solving} and Olympiad-Bench~\citep{he2024olympiadbench}. The Qwen3-8B/14B scale-up runs retain DUET's advantage on six of seven directional benchmarks. As DUET is presented as a synchronous method, we tested asynchronous method (i.e., APRIL~\citep{zhou2025april}) in Table~\ref{tab:april}. We also compared other heuristics methods include GRESO~\citep{zheng2025greso} to test whether DUET's advantage extends beyond its joint allocator--abort controller. The heuristic comparison in Table~\ref{tab:heuristic} indicates that replacing gradient variance with verifier accuracy does not consistently improve the quality--speed trade-off. These checks preserve the same quality--speed conclusion. The detailed trajectories, scale-up rows, and heuristic controls appear in Appendix~\ref{supp:robustness} (Tables~\ref{tab:trajectory}, \ref{tab:convergence-summary}, \ref{tab:scale-up}--\ref{tab:scale-tokens}, and \ref{tab:heuristic}).}
\begin{table}[t]
\caption{Sensitivity to the abort floor $\varepsilon_{\text{abort}}$ on Qwen3-1.7B / MATH at $50\%$ rollout budget. All cells share the same controller, surrogate, and seed; only $\varepsilon_{\text{abort}}$ varies.}
\label{tab:eps-sweep}
\centering
\small
\begin{tabular}{lcccccc}
\toprule
$\varepsilon_{\text{abort}}$ & MATH-500 & GSM8K & AIME-24 & HumanEval & GPQA-D & Speedup \\
\midrule
$0.01$           & 61.4 & 83.4 & \underline{12.5} & \underline{51.1} & 27.1 & \textbf{3.13$\times$} \\
$0.05$ (default) & \textbf{62.1} & \textbf{84.0} & 10.8 & \textbf{51.7} & \underline{29.1} & \underline{2.51$\times$} \\
$0.10$           & \underline{62.0} & 83.1 & \textbf{13.3} & 50.6 & 27.5 & 2.05$\times$ \\
$0.20$           & 61.9 & \underline{83.9} & \underline{12.5} & \underline{51.1} & \textbf{30.5} & 1.94$\times$ \\
\bottomrule
\end{tabular}
\end{table}

\paragraph{The performance is set by the framework, not by hyperparameter tuning.}
Table~\ref{tab:eps-sweep} sweeps $\varepsilon_{\text{abort}} \in \{0.01, 0.05, 0.10, 0.20\}$ across two orders of magnitude. MATH-500 accuracy moves only $0.7\%$ ($61.4 \to 62.1$), well inside the single-seed sampling band, and the rest of the benchmarks stay within roughly $1.4\%$ end to end. Wall-clock cost moves monotonically: $\varepsilon = 0.01$ runs at $3.13\times$ vs full-budget GRPO and $\varepsilon = 0.20$ at $1.94\times$, with the default $\varepsilon = 0.05$ at the knee of the curve ($2.51\times$). $\varepsilon_{\text{abort}}$ therefore trades wall-clock for a small SNIPS-overhead margin; the quality of the trained model is determined by the cost-weighted Neyman allocation and the marker-gated abort, not by the value of $\varepsilon_{\text{abort}}$. \rebuttaledit{We also investigate the influence of $\delta_{\rm poll}$ (marker polling interval), $W$ (detector prefix window), $s_{\rm floor}$ (minimum surrogate score), and the marker patterns (answer detector). Across these implementation choices, the advantage of DUET in its quality and speed, and the privilege of its allocator--abort design remain unchanged. This indicates that the default configuration is representative rather than a narrowly tuned point (Tables~\ref{tab:poll-window}--\ref{tab:marker-pattern} in Appendix \ref{supp:robustness}).} Overall, the result is not driven by a narrow hyperparameter choice.

\subsection{Emergent Online Properties of the Controller}\label{sec:emergent}
\begin{figure}[t]
\centering
\includegraphics[width=0.9\linewidth]{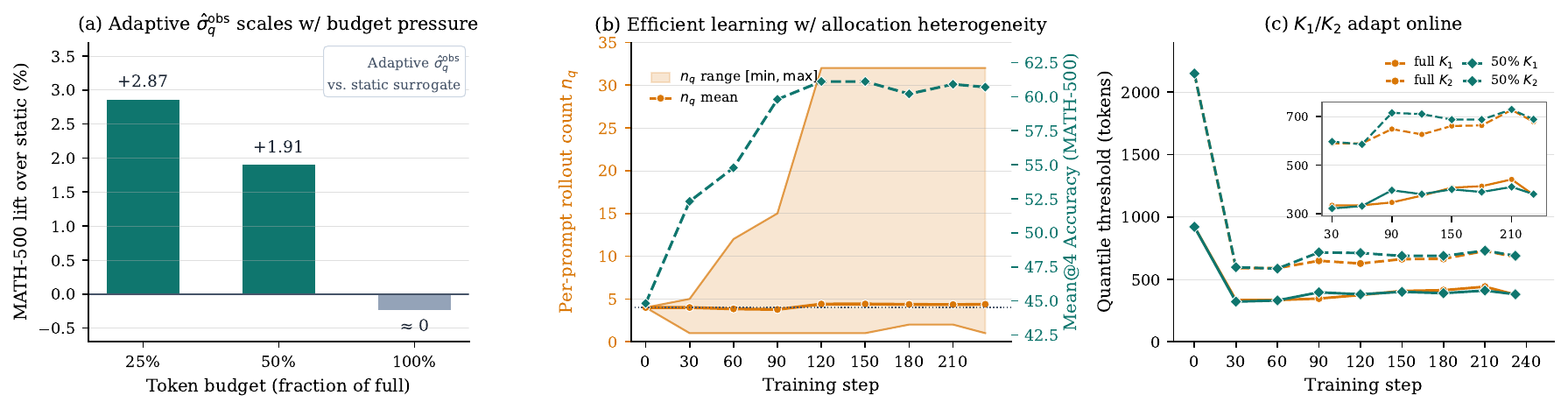}
\caption{DUET's emergent properties on Qwen3-1.7B / MATH. \textbf{(a)} Adaptive-surrogate gain over the cold-start static surrogate scales monotonically with budget pressure. \textbf{(b)} Allocation heterogeneity emerges online (with DUET b=50\%, $\varepsilon_{abort}$=0.01): the per-prompt rollout count $n_q$ starts uniform at $4$ (shaded band collapses to a point) and fans out to a range as wide as $[1, 32]$ as the online surrogate accumulates observations; MATH-500 mean@4 (right axis, dashed teal) lifts in lockstep with the fan-out, with the steepest accuracy gain coinciding with the widening of the allocation distribution. \textbf{(c)} The abort thresholds $K_1$ and $K_2$ adapt online without manual scheduling: both collapse from cold-start defaults to the policy's realised $p_{30}$ / $p_{80}$ within $30$ training steps, then track the natural-stopping length distribution as the policy's output tightens. Same trajectory at full budget (amber) and $50\%$ budget (teal).}
\label{fig:duet-dynamics}
\end{figure}
A method designed for efficiency need not display anything beyond a flat speedup. DUET, in our runs, surfaces three online behaviours that frame the controller as more than a budget knob (Figure~\ref{fig:duet-dynamics}).

\indent\xspace\xspace\xspace\xspace(1) \emph{The adaptive surrogate's value scales with budget pressure.} Replacing the cold-start static surrogate with the online running estimate $\hat\sigma^{\text{obs}}_q$ lifts MATH-500 by $+2.87\%$ at quarter budget, $+1.91\%$ at half budget, and ties at full budget (Figure~\ref{fig:duet-dynamics}(a)); the same monotone pattern holds on GSM8K (Appendix~\ref{supp:robustness}).\\
\indent\xspace\xspace\xspace\xspace(2) \emph{Heterogeneous allocation is discovered, not engineered.} Starting from $n_q = 4$ uniform across the batch, the per-prompt count fans out to a range as wide as $[1, 32]$ within one epoch (Figure~\ref{fig:duet-dynamics}(b)); the fan-out coincides with the steepest segment of the validation-accuracy curve, so the controller's internal difficulty estimate drives the model's external improvement.\\
\indent\xspace\xspace\xspace\xspace(3) \emph{The abort thresholds adapt to the policy without manual scheduling.} The early-skip $K_1$ and abort $K_2$ start from cold-start defaults near $921$ and $2150$ tokens and collapse to the policy's realised $p_{30}$ and $p_{80}$ within $30$ training steps, tracking the natural-stopping length distribution as it tightens (Figure~\ref{fig:duet-dynamics}(c)). Two kinds of token waste are eliminated as a consequence: rollouts that drift past $K_2$ without producing a parsable answer are aborted, and rollouts that have already produced one are trimmed shortly after the marker fires. The same convergence shape holds at full and half budget, so the controller is not budget-tuned.

A closer look at what the controller spends its compute on (Figure~\ref{fig:duet-internals}) exposes two additional findings.

\begingroup
\setlength{\intextsep}{10pt}
\begin{figure}[H]
\centering
\includegraphics[width=0.9\linewidth]{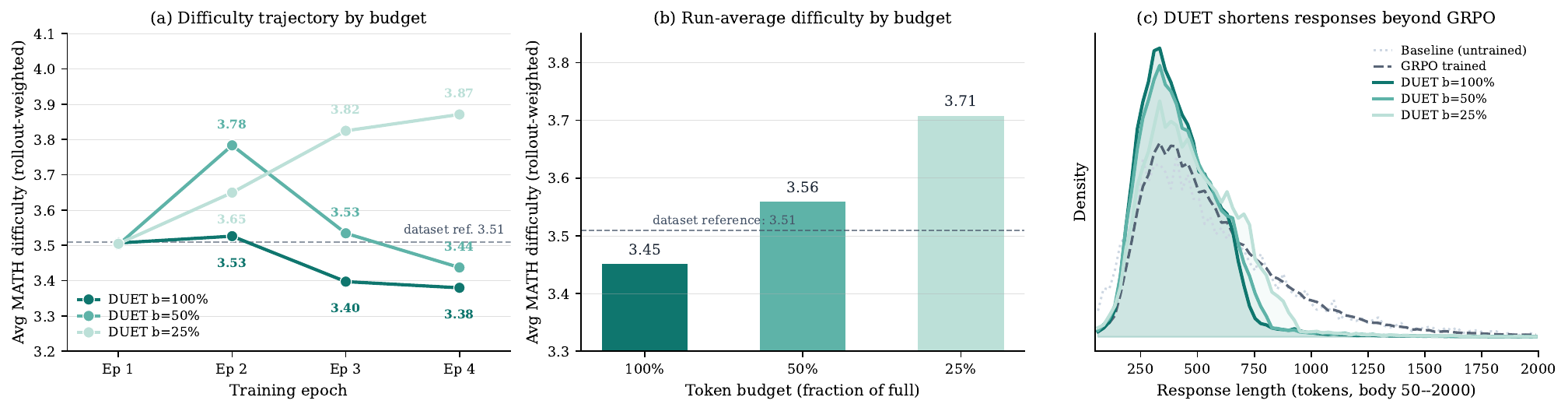}
\caption{DUET's allocation internals on Qwen3-1.7B / MATH. \textbf{(a)} Per-epoch trajectory of rollout-weighted average difficulty at three budgets, against the dataset reference $3.51$ (dashed). Full-budget DUET drifts \emph{below} the reference ($3.38$); half-budget DUET concentrates sharply at epoch~2 ($3.78$) then relaxes back below the reference ($3.44$); only at quarter budget is the concentration sustained, climbing monotonically to $3.87$. \textbf{(b)} Run-average difficulty across budgets. Tighter budget shifts the run average up. \textbf{(c)} Response-length density across DUET budgets, trained GRPO, and the untrained baseline. The baseline $\to$ GRPO shift is small (median $505 \to 511$, mean $735 \to 665$); GRPO $\to$ DUET shift is large (median $511 \to 330$).}
\label{fig:duet-internals}
\end{figure}
\endgroup

\indent\xspace\xspace\xspace\xspace(1) \emph{Difficulty concentration is jointly shaped by budget pressure and training stage.} Across the run, the rollout-weighted average difficulty rises monotonically as the budget tightens: $3.45$ at $100\%$, $3.56$ at $50\%$, $3.71$ at $25\%$, against a dataset reference of $3.51$ (Figure~\ref{fig:duet-internals}(b)). The per-epoch trajectories tell the deeper story (Figure~\ref{fig:duet-internals}(a)). At full budget the controller has enough rollouts that it never needs to concentrate: average difficulty drifts from $3.51$ at epoch~1 down to $3.38$ at epoch~4, ending below the dataset reference. At half budget the controller concentrates sharply during the steepest learning segment ($3.78$ at epoch~2, the highest point in the run) and then relaxes back to $3.44$ by epoch~4, also dipping below the dataset reference once the policy has absorbed the early hard-prompt signal. Only at quarter budget does the pressure force sustained concentration: difficulty climbs monotonically to $3.87$ by epoch~4. The picture is of a controller that spends extra rollouts on hard prompts when the budget forces it to, and surrenders that concentration the moment the budget allows.\\
\indent\xspace\xspace\xspace\xspace(2) \emph{DUET also shifts the response-length distribution toward shorter rollouts, and the shift is method-driven rather than a generic RLVR effect} (Figure~\ref{fig:duet-internals}(c)). Trained GRPO (full $232$-step run) sits almost on top of the untrained baseline (median $505 \to 511$ tokens, mean $735 \to 665$); the same RLVR loop with DUET on top collapses the median to roughly $330$ tokens and removes most of the right tail beyond $750$. The cost-weighted Neyman correction in Theorem~\ref{thm:t1} favours short, well-discriminated prompts in the per-prompt allocation, and the marker-gated abort prevents long marker-less rollouts from running to the maximum decode length; together they account for the shift.

\indent\xspace\xspace\xspace\xspace(3) \emph{\rebuttaledit{Learning to produce answers keeps the importance-correction cost small.}}
\rebuttaledit{The controller's adaptation also appears in its importance correction. Marker-less rollouts that are retained receive large weights, which can increase gradient variance. As the policy learns to produce parsable answers, fewer rollouts need this correction. In our runs, the realised second-moment factor stays near $1\times$ and never exceeds $1.5\times$ at $\varepsilon_{\text{abort}}=0.01$ (Figure~\ref{fig:theory-empirics}(c)), well below the worst-case allowance in Theorem~\ref{thm:t2}. Theorem~\ref{thm:t6} in Appendix~\ref{supp:self-extinguishing} gives an idealised counterpart: the surcharge vanishes when the policy converges to a realisable solution and the abort threshold has stabilised. Under KL regularisation, the correction remains small but non-zero in our runs.}

\section{Discussion and Limitations}\label{sec:discussion}
\rebuttaledit{RLVR learns from repeated attempts to solve a prompt, and generating these attempts accounts for much of its training cost. This could be wasteful when models are learning on signal-less prompts repeatedly. We studied whether that compute could be used more effectively by dynamically allocating prompt rollouts and response length together. DUET coordinates these decisions under a shared token budget and adapts them as the policy learns. We conducted comprehensive experiments, showing that DUET not only converges faster but also higher. It outperformed GRPO and other efficient RLVR methods with half of token budge assigned and this advantage is consistently proven by sufficient ablations on each hyperparameters.}

\rebuttaledit{Moreover, DUET's emergent behavior suggests a form of curriculum learning under a token budget. In Figure~\ref{fig:duet-dynamics}, rollout counts differentiate across prompts as validation performance improves, and learning the allocation signal matters more when compute is scarce. Figure~\ref{fig:duet-internals}(a,b) reveals the resulting emphasis, where increasingly difficult questions receive a larger share of rollouts as training progresses at quarter budget. At half budget, the concentration on harder questions later relaxes. This connects to findings from large-scale reasoning training. Kimi k1.5 reports gains from concentrating on hard questions after an initial mixed-difficulty phase~\citep{team2025kimi}, while Qwen3 selects questions that are challenging but still learnable for its starting policy~\citep{qwen2025qwen3}. These indicate that a question's training value depends on what the model can learn from it at that stage. DUET develops this changing emphasis without a prescribed difficulty schedule, with the available budget shaping how long the concentration is sustained. The shorter response distribution in Figure~\ref{fig:duet-internals}(c) adds a complementary implication: concentrating learning on difficult prompts can coexist with reducing generation overall. Together with the adaptive stopping behavior in Figure~\ref{fig:duet-dynamics}(c), this suggests a way to reserve compute for productive exploration. The savings could be reinvested in additional attempts or longer useful reasoning traces on problems that warrant them, making deeper reasoning on hard problems more affordable within a limited compute budget.}

\paragraph{Scope and limitations.}
\rebuttaledit{The current marker-gated abort assumes a machine-verifiable answer format as stated in RLVR, so our experiments only cover math, code, and short-form QA rather than subjective open-ended evaluation. Given this, extending the controller to open-ended tasks with reliable evaluators is a natural next step with our current variance-based allocator setting, where the variance-based allocator could be paired with an outcome signal suited to the evaluator. Additionally, DUET's performance may be upper-bounded by the maximum amount of rollouts $n_{max}$ allocated to each prompt. However, while computing the full ($n_{max}$) rollouts would theoretically provide the strongest learning signal, it is impractical because it requires substantially more computation (e.g., ($n_{max}$=$32$) requires roughly four times the tokens of the full-budget GRPO baseline in our main study). DUET instead targets the realistic setting of a fixed compute budget, reallocating rollouts in a variance-optimal manner, demonstrating that smarter allocation of a limited budget is more effective than uniformly increasing computation. We should study what should be the best upper bound and lower bound in such dynamic allocation process in the future. Moreover, DUET is a synchronous efficient RL method which could work together with asynchronous execution to gain the advantage from both sides. }

\paragraph{Broader impacts.}
RLVR is among the more compute-intensive stages of modern LLM post-training, with rollout generation dominating each step's cost; large-network training emissions are a material part of the field's environmental footprint~\citep{strubell2019energy,patterson2021carbon,schwartz2020green}. Methods like DUET that cut per-step token cost without giving back accuracy can lower the energy and carbon cost of future reasoning-model training, especially in the tight-budget regime relevant to academic and resource-limited labs. We do not foresee additional misuse risks specific to DUET beyond those already inherent in RLVR-trained reasoning models.

\rebuttaledit{\section{Acknowledgement}
The authors are supported by NSF  grant 2523500. Additionally, the author was supported by the \href{https://modal.com/academics}{Modal Academic Compute Grant} for project exploration and part of experiments. The author also want to thank \href{https://tinker-docs.thinkingmachines.ai/tinker/}{Tinker}'s support to Columbia University on computing credits for project exploration.}


\small
\bibliographystyle{plainnat}
\bibliography{references}

\appendix
\newpage

\section{Appendix}\label{supp:appendix}{\color{black}\begin{center}\Large\textbf{Appendix Contents}\end{center}\begingroup\Large\setlength{\parskip}{1.1em}\noindent\hyperref[supp:related]{A.1\quad Related Work}\dotfill\pageref{supp:related}\par\noindent\hyperref[supp:duet-formal]{A.2\quad The Theory Supplement of DUET}\dotfill\pageref{supp:duet-formal}\par\noindent\hyperref[supp:experimental-setup]{A.3\quad Detailed Experimental Setup}\dotfill\pageref{supp:experimental-setup}\par\noindent\hyperref[supp:theory-empirics]{A.4\quad Theory--Empirics Alignment}\dotfill\pageref{supp:theory-empirics}\par\noindent\hyperref[supp:robustness]{A.5\quad Ablation Studies: Robustness and Sensitivity}\dotfill\pageref{supp:robustness}\par\noindent\hyperref[supp:cross-scale]{A.6\quad Cross-Scale and Cross-Family Detail}\dotfill\pageref{supp:cross-scale}\par\noindent\hyperref[supp:quarter-arrol]{A.7\quad Quarter-Budget Panel and ARRoL Faithful Detail}\dotfill\pageref{supp:quarter-arrol}\par\noindent\hyperref[supp:prompts]{A.8\quad Prompt Templates}\dotfill\pageref{supp:prompts}\par\noindent\hyperref[supp:case-studies]{A.9\quad Case Studies: Baseline-Model Output vs DUET-Trained Output}\dotfill\pageref{supp:case-studies}\par\noindent\hyperref[supp:algorithm]{A.10\quad DUET Algorithm}\dotfill\pageref{supp:algorithm}\par\noindent\hyperref[supp:llm-use]{A.11\quad The Use of LLM}\dotfill\pageref{supp:llm-use}\par\endgroup}
\clearpage

\subsection{Related Work}\label{supp:related}

\paragraph{Budget-Constrained Policy-Gradient Estimation.}
Stratified importance sampling with per-stratum cost is widely-used as the standard~\citep{owen2013monte}, in which the variance-minimizing allocation is the cost-weighted Neyman optimum $n_k^\star \propto \sigma_k/\sqrt{c_k}$. \citet{carpentier2012adaptive} give the canonical adaptive-stratification bandit with sub-linear regret to the oracle. VIP~\citep{nguyen2026vip} first instantiated the count-axis Neyman optimum in RLVR with a Gaussian-process surrogate over prompt embeddings; DynaMO~\citep{fang2026dynamo} concurrently derived the count-axis allocation with a Bernoulli-variance proxy and added token-level advantage modulation. Both treat rollout length as exogenous and minimize variance under a count budget. We instead minimize variance under a \emph{token} budget, which adds the $\sqrt{\bar L_q}$ factor to the prompt allocation and couples it to a within-rollout abort rule through a shared budget constraint. The single Lagrange multiplier $\lambda^\star$ closes the budget given the abort policy; the abort threshold $K_2$ is a quantile heuristic, not a dual minimiser.

\paragraph{Selective Generation in RLVR.}
Three RLVR methods cut wasted computation along distinct axes. GRESO~\citep{zheng2025greso} skips a prompt whose history of group rewards has been degenerate, using a moving-window threshold; the rule is binary and tuned per dataset. DAPO~\citep{yu2025dapo} keeps the uniform rollout fan-out and post-filters degenerate groups, paying the generation cost on the dropped samples and offering no off-policy correction for the induced bias. ARRoL~\citep{xu2026arrol} installs a quality head at a fixed inspection point and Bernoulli-prunes mid-decode. SEED-GRPO~\citep{chen2025seedgrpo} re-weights groups by semantic-entropy uncertainty without changing the rollout count or length and is therefore complementary. None of these methods coordinates the three control axes (prompt, count, length) under a single optimization, takes a hard token budget as the optimization target, or carries a self-correcting safety property whereby the cost of bias-cancellation reweighting decays as the policy improves.

\paragraph{Off-Policy Evaluation under Adaptive Sampling.}
The self-normalized estimator~\citep{swaminathan2015}, doubly robust evaluation~\citep{jiang2016dr}, and weighted doubly robust estimators~\citep{thomas2016wdr} supply the unbiasedness machinery for evaluating a policy under a sampling rule different from the one that generated the data. \citet{liu2018curse} extend importance sampling to infinite-horizon stationary-distribution targets. None of these works addresses the joint multi-level adaptive selection that arises when the prompt-level decision (how many rollouts) and the within-rollout decision (when to abort) are both made by a learned controller. Our combination of a per-prompt stratification weight $s^{\text{pre}}$ with an abort-gate IS correction $1/p^{\text{abort}}$ and an $\varepsilon$-floor on the abort coin is, to our knowledge, the first such design in RLVR; we analyse its per-rollout unbiasedness under an action-independent baseline.

\paragraph{Mid-Decode Termination in LLM Serving.}
vLLM logits processors~\citep{kwon2023pagedattention}, speculative decoding~\citep{leviathan2023fast}, and the EAGLE family of speculative methods all manipulate the decode loop to cut inference latency. We use the same per-prompt logits-processor machinery but instantiate a \emph{training-time} termination rule with provable unbiasedness. The abort decision uses a domain-specific marker detector polled every $\delta_{\text{poll}}$ tokens against a bounded prefix window; $\varepsilon$-keep, IS reweighting on the abort gate, and the training-side gradient mask together preserve per-rollout unbiasedness for the action-independent-baseline estimator (Theorem~\ref{thm:t2}). Inference-time speed methods do not need this correction and do not provide it; using them as-is during training would bias the gradient.

\paragraph{Length Penalties and Truncation Failures.}
A growing line of work documents the failure modes of fixed-length truncation in RLVR. Xiang et al.~\citep{xiang2025justenough} show that fixed caps inject reward noise (coherent but unfinished traces are penalised identically to wrong answers) and push the policy toward premature low-effort outputs; their adaptive length penalty, scaled inversely to per-prompt online solve rate, halves response length on DeepScaleR-1.5B without accuracy loss. Yuan et al.~\citep{yuan2025lazypenalty} apply a length penalty only after the policy is producing correct answers, cutting response length by $33$--$40\%$ at parity. DUET takes a different route: rather than penalising long responses, the marker-gated abort terminates only the rollouts that have failed to produce a parsable answer past a quantile of the policy's own natural-stopping distribution, leaving the policy free to use as many tokens as it needs to reach an answer.

\paragraph{Test-Time Compute Scaling.}
The test-time compute literature uses the same budget convention DUET adopts: measure inference allocation in \emph{generated tokens}, not FLOPs. Snell et al.~\citep{snell2024scaling} parameterise compute as $N$ samples per query at deploy time; the s1 work~\citep{muennighoff2025s1} writes explicitly that ``tokens generated correspond to the amount of test-time compute spent.'' DUET applies the same measurement to RL rollout generation. The justification is the same: per-token FLOPs is a constant for a fixed model, autoregressive decoding is memory-bandwidth-bound rather than arithmetic-bound~\citep{leviathan2023fast,kwon2023pagedattention}, and generated-token count is the better wall-clock proxy. The pre-training scaling laws of Kaplan~\citep{kaplan2020scaling} and Chinchilla~\citep{hoffmann2022training} still use FLOPs as the canonical unit because pre-training varies architecture and dataset size; in the RL phase those degrees of freedom close, leaving generated tokens as the binding axis.

\paragraph{Curated Multi-Pass RLVR.}
DUET assumes a finite curated training corpus traversed for $E\ge 2$ epochs, the regime that dominates published RLVR practice. DAPO trains on $17$k curated prompts~\citep{yu2025dapo}; SimpleRL-Zoo on roughly $8$k~\citep{zeng2025simplerlzoo}; Light-R1 explicitly runs three RL epochs on a $3$k curated set~\citep{wen2025lightr1}; Open-Reasoner-Zero passes its $129$k pool more than once before re-training on a $13$k hard-mined subset~\citep{hu2025orz}. The ``less is more'' line~\citep{li2025limr,ye2025limo,wang2025oneshotrlvr} makes the case more directly: LIMR finds $1.4$k carefully-chosen prompts from the same MATH source we use match or beat training on the full $8.5$k pool, and one-shot RLVR lifts MATH-500 from $36\%$ to $73.6\%$ with a single training example revisited thousands of times. The binding constraint in RLVR is therefore not raw prompt count but how many times each useful prompt is \emph{visited}, which is precisely the regime the per-prompt running-mean surrogate exploits.

\paragraph{Best-Arm Identification and Budgeted Simulation.}
The classical lineage that DUET inherits comes from operations research and bandits. Best-arm identification under fixed budgets~\citep{evendar2006action,audibert2010best} couples the allocation question to a stopping rule for each arm. Optimal-budget allocations like OCBA~\citep{chen2000ocba} use one posterior-derived score to govern both how many simulations a candidate receives and when its run can be retired, and budget-constrained simulation~\citep{glynn1992asymptotic} treats the budget as the primary object and derives allocation from a cost-shadowed efficiency criterion. DUET imports the OCBA-style joint-control move into RLVR: one shared budget constraint couples the prompt-level allocation (variance-optimal under cost-weighted Neyman, Theorem~\ref{thm:t1}) with the within-rollout abort (a marker-gated quantile heuristic, analysed for unbiasedness in Theorem~\ref{thm:t2}), instead of treating the two decisions as independent levers as the prior RLVR efficiency literature does.

\paragraph{Capability Comparison.}
Table~\ref{tab:capabilities} consolidates the contrast on capabilities relevant to any budget-aware online RLVR method. Rows are method-neutral so the same column applies to future baselines. Cells use \checkmark = full, $\circ$ = partial / single-axis / post-hoc, $-$ = absent.

\begin{table}[h]
  \caption{Capability comparison across budget-aware RLVR methods. The diagonal pattern of \checkmark marks among the baselines reflects the single-axis design of each: DAPO acts at the group level post-rollout; ARRoL acts within rollouts; VIP acts across prompts. DUET coordinates all three axes under a single shared token budget. $\dagger$ = result holds in the idealised theorem setting (action-independent baseline, true lengths $\bar L_q$, fixed $K_2$); the practical implementation uses GRPO group normalisation, estimated lengths, and integer rounding.}
  \label{tab:capabilities}
  \centering
  \small
  \begin{tabular}{lccccc}
    \toprule
    Capability & GRPO & DAPO & ARRoL & VIP & \textbf{DUET} \\
    \midrule
    Adaptive prompt selection           & $-$ & $\circ$ & $-$ & \checkmark & \checkmark \\
    Variable rollout count per prompt   & $-$ & $-$ & $-$ & \checkmark & \checkmark \\
    Adaptive within-rollout length      & $-$ & $-$ & \checkmark & $-$ & \checkmark \\
    Multi-axis joint coordination       & $-$ & $-$ & $-$ & $-$ & \checkmark \\
    Pre-rollout decision (saves generation) & $-$ & $-$ & $-$ & \checkmark & \checkmark \\
    Explicit compute-budget constraint  & $-$ & $-$ & $\circ$ & $\circ$ & \checkmark \\
    Bias-corrected gradient under selection & \checkmark & $\circ$ & $\circ$ & \checkmark & $\circ^\dagger$ \\
    Variance-optimality guarantee       & $-$ & $-$ & $-$ & $\circ$ & $\circ^\dagger$ \\
    Self-correcting safety overhead     & $-$ & $-$ & $-$ & $-$ & $\circ^\dagger$ \\
    \bottomrule
  \end{tabular}
\end{table}

\subsection{Theory Supplement of DUET}\label{supp:duet-formal}

This subsection collects the notation, assumptions, formal statements of Theorems~\ref{thm:t4} and \ref{thm:t2} (deferred from Section~\ref{sec:method}), proofs of all four theorems and Proposition~\ref{prop:t3}, and Algorithm~\ref{alg:duet} (the full DUET training step). Theorem~\ref{thm:t1} is stated and used inline in Section~\ref{sec:method}; Theorem~\ref{thm:t6} appears below for the first time.

\subsubsection{Notation}\label{supp:notation}

{\scriptsize
\begin{tabular}{ll}
\toprule
Symbol & Meaning \\
\midrule
$\pi_\theta$ & Actor policy, $\theta\in\mathbb{R}^d$ \\
$\mathcal{B}_t \subset \mathcal{Q}$ & Prompt batch at step $t$, $|\mathcal{B}_t|=M$ \\
$L_{q,i}$, $\bar L_q(\theta)$ & Random length / realized expected length of rollout $i$ at prompt $q$ \\
$R(q,y)$ & Verifier reward, $|R|\le R_{\max}$ \\
$n_q$, $n_q^\star$ & Allocated / Neyman-optimal rollout count for prompt $q$ \\
$A_{q,i}$, $Z_{q,i}$ & GRPO advantage; per-rollout gradient contribution $A_{q,i}\nabla_\theta\log\pi_\theta(y_{q,i}\mid q)$ \\
$\sigma_q^2$ & Trace within-prompt covariance of $Z_{q,i}$ \\
$V(n)$, $V^\star$ & Trace variance of $\hat g_t$; minimum at Neyman optimum \\
$B$, $\lambda^\star$ & Token budget per step; budget-closing dual multiplier \\
$S$ & Neyman normalizer $\sum_k \sigma_k\sqrt{\bar L_k}$ \\
$\hat s_q$ & Surrogate for $\sigma_q$ \\
$\chi^2(\hat s,\sigma)$, $\bar\chi^2$ & Calibration divergence; time-averaged version \\
$m(y^{1:t})$ & Domain-specific marker indicator \\
$\tau_q^{\text{marker}}$, $\tau_q^{\text{abort}}$ & Marker-emission time; abort time \\
$K_1, K_2$ & Online $p_{30}, p_{80}$ quantiles of natural-stopping lengths \\
$G$, $W$, $\delta_{\text{poll}}$ & Grace window; sliding-window size; marker poll interval \\
$I_{q,i}^{\text{abort}}$, $p_{q,i}^{\text{abort}}$ & Abort-and-mask indicator; kept-rollout propensity \\
$\varepsilon_{\text{pre}}, \varepsilon_{\text{abort}}$ & Lower clip on $s^{\text{pre}}$; abort exploration floor \\
$p_q^{\text{marker}}(\theta)$, $p_{\text{marker}}(\theta)$ & Per-prompt / batch-mean marker-emission rate \\
$\sigma^2_{\text{m-less},q}$ & Conditional second moment $\mathbb{E}[\|Z_{q,i}^{\text{nat}}\|^2\mid \tau_q^{\text{marker}} > K_2 + G]$ (no marker by the abort gate) \\
$\eta$, $\mu$ & Step size; local strong-concavity constant (decay-rate corollary only) \\
$G_{\max}$ & Bound on $\|\nabla_\theta\log\pi_\theta\|$ \\
\bottomrule
\end{tabular}}

\subsubsection{Assumptions}\label{supp:assumptions}

\paragraph{(A2) Bounded log-gradient.} $\|\nabla_\theta\log\pi_\theta(y\mid q)\|\le G_{\max}$ for every encountered $(q,y,\theta)$.

\paragraph{(A3) Bounded reward and marker-prefix sufficiency.} $|R(q,y)|\le R_{\max}$. For marker-emitting rollouts, $R(q,y)$ is determined by the prefix up to $\tau_q^{\text{marker}}$ (the marker detector fires only when the answer span is complete, so the verifier's reading of $y$ is fixed at that point). Any stop time $\tau\ge \tau_q^{\text{marker}}$ therefore preserves $R$. $\varepsilon$-kept marker-less rollouts are scored at natural EOS as usual.

\paragraph{(A5) Abort exploration floor.} Every marker-less rollout is kept with probability $\varepsilon_{\text{abort}} > 0$. The pre-rollout weight $s^{\text{pre}}_q$ is a stratification factor used to balance per-prompt aggregation; no actual prompt-level sampling occurs.

\paragraph{(A6) Pointwise variance bounds.} There exist $\sigma_{\text{low}}, \sigma_{\text{high}} > 0$ with $\sigma_{\text{low}} \le \sigma_q \le \sigma_{\text{high}}$ for every $q\in\mathcal{Q}$, and prompt mass $\mu(q) \ge \mu_{\min} > 0$. Theorem~\ref{thm:t4} therefore applies on the regular subset of prompts; degenerate zero-variance groups (e.g., GRPO groups in which all rollouts receive identical reward) are excluded from this analysis, and the implementation handles them by a small additive regularisation $\sqrt{\sigma_q^2 + \epsilon^2}$ in the std normaliser. The surrogate cold-start floor $s_{\text{floor}}$ used in $\hat s_q := \max\{s_{\text{floor}}, \hat\sigma_q^{\text{obs}}\}$ is a separate quantity (chosen as the 5th percentile of $\hat\sigma_q^{\text{obs}}$ after epoch~1 in our experiments).

\paragraph{(A6$'$) Marker continuity, solvability, marker horizon, and realizability.} Throughout, $K_2$ is held fixed (or has stabilised to a steady-state value). (i) The map $\theta\mapsto p_{\text{marker}}(\theta)$ is continuous on $\{\theta_t\}_{t\le T}$. (ii) Every prompt $q\in\mathcal{Q}$ admits some trajectory $y$ with $R(q,y)>0$ and $m(y^{1:t})=1$ for some $t\le K_2+G$. (iii) The verifier-optimal limit $\theta^\star$ realizes a policy that places all probability on trajectories of type (ii) on every prompt; equivalently, the policy class is rich enough to express such an optimum without regularization-induced spread.

\subsubsection{Proof of Theorem~\ref{thm:t1}}\label{supp:t1-proof}

Each term $\sigma_q^2/n_q$ is strictly convex in $n_q$ on $(0,\infty)$ (second derivative $2\sigma_q^2/n_q^3 > 0$), so $V$ is strictly convex on $\mathbb{R}_{>0}^M$. The feasible set $\{n\in\mathbb{R}_{>0}^M : \sum n_q\bar L_q = B\}$ is a convex affine slice, so any KKT stationary point is the unique minimizer.

Introduce the Lagrangian $\mathcal{L}(n,\lambda) = \sum_q\sigma_q^2/n_q + \lambda(\sum_q n_q\bar L_q - B)$. Stationarity in $n_q$:
$$\partial_{n_q}\mathcal{L} = -\sigma_q^2/n_q^2 + \lambda\bar L_q = 0,$$
giving $n_q = \sigma_q/\sqrt{\lambda\bar L_q}$ with $\lambda>0$ for positivity. Substituting into the budget,
$$B = \sum_q n_q\bar L_q = \frac{1}{\sqrt\lambda}\sum_q\sigma_q\sqrt{\bar L_q} = \frac{S}{\sqrt\lambda},$$
so $\sqrt\lambda = S/B$. Back-substituting yields $n_q^\star = (B/S)\sigma_q/\sqrt{\bar L_q}$ and $V^\star = S^2/B$. Strict convexity gives uniqueness. $\square$

\paragraph{Corollary (uniform comparison).} At matched budget, $V_{\text{unif}}/V^\star = (\sum_q\sigma_q^2)(\sum_k\bar L_k)/(\sum_q\sigma_q\sqrt{\bar L_q})^2 \ge 1$ by Cauchy--Schwarz, with equality iff $\sigma_q/\sqrt{\bar L_q}$ is constant across prompts.

\paragraph{Remark on integer rounding.} The implemented allocation rounds each $n_q^\star$ to a positive integer and clips at $n_{\min}$. When the continuous optimum satisfies $n_q^\star \ge n_{\min}$ for every prompt and $B \gg M\bar L_{\max}$, rounding perturbs the budget by at most $O(M\bar L_{\max})$ tokens and the variance excess is $O(M\bar L_{\max}/B)$. At very tight budgets where $n_{\min}$ is active, clipping dominates and the bound no longer applies.

\subsubsection{Proposition: Joint-Controller Budget Feasibility}\label{supp:t3-proof}

\begin{proposition}[Joint-controller budget feasibility]\label{prop:t3}
Fix the abort policy of Section~\ref{sec:method-abort} with floor $\varepsilon_{\text{abort}}$ and grace $G$, and let $\bar L_q(\theta)$ be the realised expected length under $\pi_\theta$ and that abort policy. Then there exists a unique $\lambda^\star(\theta) > 0$ such that the cost-weighted Neyman allocation $n_q^\star = \sigma_q/\sqrt{\lambda^\star(\theta)\,\bar L_q(\theta)}$ satisfies $\sum_q n_q^\star\,\bar L_q(\theta) = B$.
\end{proposition}

\paragraph{Proof.} Fix $\theta$. The realised $\bar L_q(\theta)$ is a constant in $\lambda$, so $n_q(\lambda,\theta) = \sigma_q/\sqrt{\lambda\bar L_q(\theta)}$ is continuous and strictly decreasing in $\lambda$, with $n_q(0^+) = \infty$ and $n_q(\infty) = 0$. Therefore
$$\Phi_\theta(\lambda) \;:=\; \sum_q n_q(\lambda,\theta)\bar L_q(\theta) \;=\; \frac{S(\theta)}{\sqrt\lambda},\qquad S(\theta) := \sum_q \sigma_q\sqrt{\bar L_q(\theta)},$$
is continuous and strictly decreasing on $(0,\infty)$ with $\Phi_\theta(0^+)=+\infty$ and $\Phi_\theta(\infty)=0$. By the intermediate value theorem there is a unique $\lambda^\star(\theta)$ with $\Phi_\theta(\lambda^\star)=B$. Substituting into Theorem~\ref{thm:t1}'s stationarity condition recovers the cost-weighted Neyman form. $\square$

\paragraph{Remark on $K_2$.} The threshold $K_2$ is the $p_{80}$ quantile of natural-stopping lengths under the current policy, refit online (Section~\ref{sec:method-abort}). Proposition~\ref{prop:t3} is therefore a budget-feasibility statement under the abort policy induced by the current $\theta$ and the current $K_2$; Theorem~\ref{thm:t1}'s variance optimality remains valid under the resulting $\bar L_q(\theta)$.

\subsubsection{Theorem~\ref{thm:t4}: Surrogate Calibration Bound}\label{supp:t4-proof}

Phase~(i) of DUET (Section~\ref{sec:method-alloc}) substitutes the running surrogate $\hat s_q$ for the true within-prompt standard deviation $\sigma_q$ in the cost-weighted Neyman rule of Theorem~\ref{thm:t1}. The resulting plug-in allocation $n_q^s$ no longer reaches the variance optimum $V^\star$ of Theorem~\ref{thm:t1}; the next theorem bounds the gap by a single calibration divergence.

\begin{theorem}[Surrogate calibration bound]\label{thm:t4}
Let $n_q^s$ be the surrogate plug-in of Theorem~\ref{thm:t1} with $\hat s_q$ in place of $\sigma_q$, and assume true lengths $\bar L_q$ are used in the allocation. Under pointwise variance bounds,
\begin{equation}
V(n^s) - V^\star \;\le\; K\,\frac{\chi^2(\hat s,\sigma)}{B},
\label{eq:t4}
\end{equation}
where $K$ depends only on the variance constants, length range, $s_{\text{floor}}$, and batch size $M$.
\end{theorem}

\paragraph{Proof.} Write $r_q := \hat s_q/\sigma_q$, $a_q := \sigma_q\sqrt{\bar L_q}$, so $S = \sum a_q$ and $\tilde S := \sum a_q r_q$. Then
$$V(n^s) = \sum_q\sigma_q^2\cdot\frac{\tilde S\sqrt{\bar L_q}}{B\hat s_q} = \frac{\tilde S}{B}\sum_q\frac{a_q}{r_q} = \frac{1}{B}\Bigl(\sum_q a_q r_q\Bigr)\Bigl(\sum_q\frac{a_q}{r_q}\Bigr) = \frac{S^2}{B}\,\mathbb{E}_w[r]\,\mathbb{E}_w[1/r],$$
where $w_q := a_q/S$. Since $V^\star = S^2/B$,
$$V(n^s) - V^\star = V^\star\bigl(\mathbb{E}_w[r]\,\mathbb{E}_w[1/r] - 1\bigr).$$

Define $\chi^2_w(\hat s,\sigma) := \mathbb{E}_w[(r-1)^2]$ under the weighted measure $w$, and $\chi^2(\hat s,\sigma) := \mathbb{E}_\mu[(r-1)^2]$ under the prompt distribution $\mu$. Let $r, r'$ be i.i.d.\ under $w$. The identity $\mathbb{E}_w[r]\,\mathbb{E}_w[1/r] - 1 = \tfrac{1}{2}\mathbb{E}[(r-r')^2/(rr')]$ holds by direct expansion. Since $\hat s_q\ge s_{\text{floor}}$ by construction (the surrogate cold-start floor) and $\sigma_q\le\sigma_{\text{high}}$ by (A6), $r_q\ge r_{\min} := s_{\text{floor}}/\sigma_{\text{high}} > 0$ pointwise, so $1/(rr')\le 1/r_{\min}^2$. Combined with $\mathbb{E}[(r-r')^2] = 2\operatorname{Var}_w(r) \le 2\chi^2_w(\hat s,\sigma)$,
$$\mathbb{E}_w[r]\,\mathbb{E}_w[1/r] - 1 \;\le\; \chi^2_w(\hat s,\sigma)/r_{\min}^2.$$

All sums and measures in this proof are over the per-step batch $\mathcal{B}_t$ of size $M$: $S = \sum_{q\in\mathcal{B}_t} \sigma_q\sqrt{\bar L_q}$, $w_q = a_q/S$ is supported on $\mathcal{B}_t$, and $\chi^2$ is taken under the empirical batch measure $\mu_t(q) = 1/M$ for $q\in\mathcal{B}_t$. Under (A6), $\sigma_{\text{low}} \le \sigma_q \le \sigma_{\text{high}}$ and $L_{\min}\le \bar L_q \le L_{\max}$, so $S \ge M\sigma_{\text{low}}\sqrt{L_{\min}}$. The pointwise density ratio $dw/d\mu_t = M\sigma_q\sqrt{\bar L_q}/S$ satisfies $\rho := \sup_q dw/d\mu_t \le \sigma_{\text{high}}\sqrt{L_{\max}}/(\sigma_{\text{low}}\sqrt{L_{\min}})$, depending only on the (A6) constants. Then $\chi^2_w \le \rho\,\chi^2$. Also $V^\star = S^2/B \le \sigma_{\text{high}}^2 M^2 L_{\max}/B$. Combining,
$$V(n^s) - V^\star \;\le\; \frac{V^\star\,\rho\,\chi^2(\hat s,\sigma)}{r_{\min}^2} \;\le\; K\,\frac{\chi^2(\hat s,\sigma)}{B}, \qquad K := \frac{\sigma_{\text{high}}^2 M^2 L_{\max}\,\rho}{r_{\min}^2}, $$
with $K$ a function only of $(\sigma_{\text{low}}, \sigma_{\text{high}}, L_{\min}, L_{\max}, s_{\text{floor}}, M)$ as claimed. $\square$

\paragraph{Remark on length estimation.} Theorem~\ref{thm:t4} is stated with true $\bar L_q$. The implementation uses $\hat L_q$, which enters both the per-prompt allocation $\hat s_q/\sqrt{\lambda\hat L_q}$ and the bisection that closes the budget; we conjecture an analogous calibration bound $O(\chi^2(\hat L,\bar L)/B)$ holds, but the joint analysis with surrogate lengths is left to future work.

\subsubsection{Theorem~\ref{thm:t2}: Marker-Gated Abort Unbiasedness and Variance Increase}\label{supp:t2-proof}

Phase~(ii) of DUET (Section~\ref{sec:method-abort}) keeps a marker-emitting rollout at propensity $1$ and $\varepsilon$-keeps a marker-less rollout with probability $\varepsilon_{\text{abort}}$, importance-reweighting the kept marker-less rollout by $1/\varepsilon_{\text{abort}}$. The next theorem certifies that this $\varepsilon$-keep with reweighting design preserves per-rollout unbiasedness for the action-independent-baseline gradient and bounds the second-moment surcharge it incurs.

\begin{theorem}[Marker-gated abort: per-rollout unbiasedness and variance increase]\label{thm:t2}
Under (A2)--(A3) and an action-independent baseline (so $A_{q,i}$ does not couple to other rollouts' abort coins), the per-rollout IS-corrected contribution $\hat h_{q,i} := (1 - I_{q,i}^{\text{abort}})\, Z_{q,i}\, /\, p_{q,i}^{\text{abort}}$ is unbiased for the natural-rollout contribution $Z_{q,i}^{\text{nat}}$, and its trace second moment satisfies
\begin{equation}
\mathbb{E}\!\bigl[\|\hat h_{q,i}\|^2\bigr] - \mathbb{E}\!\bigl[\|Z_{q,i}^{\text{nat}}\|^2\bigr]
\;\le\; (1 - p_q^{\text{marker}})\,\frac{1-\varepsilon_{\text{abort}}}{\varepsilon_{\text{abort}}}\,\sigma^2_{\text{m-less},q},
\label{eq:t2-variance}
\end{equation}
where $p_q^{\text{marker}} := \mathbb{P}(\tau_q^{\text{marker}} \le K_2 + G)$ is the per-prompt rate of marker emission by the abort gate and $\sigma^2_{\text{m-less},q}$ is the conditional second moment of $Z_{q,i}^{\text{nat}}$ given no marker by that gate.
\end{theorem}

\paragraph{Setup.} Fix step $t$, prompt $q$, allocation $n_q$. For each rollout $(q,i)\in[n_q]$, generate until $\min\{\tau_q^{\text{marker}}+G,\,K_2+G,\,L_{\max}\}$. If the marker fires by $\tau_q^{\text{abort}}$, the rollout is kept ($p^{\text{abort}}=1$, $I^{\text{abort}}=0$). Otherwise draw $U_{q,i}\sim\text{Uniform}[0,1]$ independently: if $U_{q,i}<\varepsilon_{\text{abort}}$ continue to natural EOS / $L_{\max}$, $\varepsilon$-kept ($p^{\text{abort}} = \varepsilon_{\text{abort}}$, $I^{\text{abort}}=0$); else terminate at $\tau_q^{\text{abort}}$, masked-out ($I^{\text{abort}} = 1$).

For marker-emitting rollouts, (A3) gives that $R$ depends only on tokens up to $\tau_q^{\text{marker}}+G$. Force-stopping at that time drops post-$\tau_q^{\text{marker}}+G$ score-function terms, each with zero conditional expectation given the prefix; the truncation does not bias $Z_{q,i}$. Marker-less rollouts are not truncated (they are either $\varepsilon$-kept and run to natural EOS, or aborted and masked out).

Let $M_{q,i}:=\mathbf{1}\{\tau_q^{\text{marker}}\le\tau_q^{\text{abort}}\}$ and let $E_{q,i}\sim\text{Bernoulli}(\varepsilon_{\text{abort}})$ be the $\varepsilon$-keep coin (independent of the rollout). Then
$$1 - I_{q,i}^{\text{abort}} \;=\; M_{q,i} + (1 - M_{q,i})E_{q,i},\qquad p_{q,i}^{\text{abort}} \;=\; M_{q,i} + (1-M_{q,i})\varepsilon_{\text{abort}}.$$

Let $Z^{\text{nat}}_{q,i}$ denote the per-rollout contribution under the natural rollout (no abort, run to natural EOS), and $Z_{m+G,q,i}$ the contribution under tail-trim at $\tau_q^{\text{marker}}+G$. The kept estimator is $\hat h = Z_{m+G}$ when $M=1$, $Z^{\text{nat}}/\varepsilon_{\text{abort}}$ when $M=0,E=1$, and $0$ when $M=0,E=0$.

\paragraph{Unbiasedness.} The keep coin $E$ is fresh randomness drawn at $K_2+G$, hence independent of the prefix $y^{1:K_2+G}$ and of $M$ (which is determined by that prefix). We condition on $E$ rather than asserting unconditional independence:
\begin{align*}
\mathbb{E}\!\left[\tfrac{1-I^{\text{abort}}}{p^{\text{abort}}}\,Z\,\Big|\,q,\theta\right]
&= \mathbb{E}[M\,Z_{m+G}\mid q,\theta] + \mathbb{P}(E{=}1)\,\mathbb{E}\!\left[(1-M)\tfrac{Z^{\text{nat}}}{\varepsilon_{\text{abort}}}\,\Big|\,q,\theta,E{=}1\right] \\
&= \mathbb{E}[M\,Z_{m+G}\mid q,\theta] + \mathbb{E}[(1-M)\,Z^{\text{nat}}\mid q,\theta,E{=}1].
\end{align*}
Conditional on $E{=}1$, the continuation past $K_2+G$ follows $\pi_\theta$, so the conditional distribution of $Z^{\text{nat}}$ given $M{=}0$ matches the natural-rollout distribution. By (A3), $Z_{m+G}$ has the same conditional mean as $Z^{\text{nat}}$ under $M{=}1$ (post-$\tau^{\text{marker}}+G$ score terms have zero conditional mean by the score-function identity). Combining,
$$\mathbb{E}[\hat h\mid q,\theta] = \mathbb{E}[M\,Z^{\text{nat}}\mid q,\theta] + \mathbb{E}[(1-M)\,Z^{\text{nat}}\mid q,\theta] = \mathbb{E}[Z^{\text{nat}}\mid q,\theta].$$
Aggregating over $i$ and $q$ gives the per-rollout unbiasedness, provided the advantage $A_{q,i}$ is constructed from an action-independent baseline (so $A_{q,i}$ does not depend on the abort coins $E_{q,j}$ for $j\ne i$). The proof does not extend to the full GRPO group-normalised advantage, where $\bar R_q$ and $\hat\sigma(R_q)$ are computed from the same group of (potentially aborted) rollouts; the residual coupling there is a known limitation that we do not formally bound, and we report it as observed empirical bias.

\paragraph{Variance increase.} Using $E^2{=}E$ since $E\in\{0,1\}$:
\begin{align*}
\mathbb{E}\!\left[\|\hat h\|^2\,\Big|\,q,\theta\right]
&= \mathbb{E}[M\,\|Z_{m+G}\|^2\mid q,\theta] + \tfrac{1}{\varepsilon_{\text{abort}}}\,\mathbb{E}[(1-M)\,\|Z^{\text{nat}}\|^2\mid q,\theta,E{=}1] \\
&= p_q^{\text{marker}}\,\mathbb{E}[\|Z_{m+G}\|^2\mid M{=}1] + \tfrac{1-p_q^{\text{marker}}}{\varepsilon_{\text{abort}}}\,\sigma^2_{\text{m-less},q}.
\end{align*}
Compare to the all-natural-EOS baseline second moment $p_q^{\text{marker}}\mathbb{E}[\|Z^{\text{nat}}\|^2\mid M{=}1] + (1-p_q^{\text{marker}})\sigma^2_{\text{m-less},q}$. Subtracting,
$$\Delta_q \;=\; \underbrace{p_q^{\text{marker}}\bigl(\mathbb{E}[\|Z_{m+G}\|^2\mid M{=}1] - \mathbb{E}[\|Z^{\text{nat}}\|^2\mid M{=}1]\bigr)}_{\le\, 0} + (1-p_q^{\text{marker}})\,\tfrac{1-\varepsilon_{\text{abort}}}{\varepsilon_{\text{abort}}}\,\sigma^2_{\text{m-less},q}.$$
The first term is non-positive: by (A3) and the score-function identity, post-$\tau^{\text{marker}}+G$ tokens have zero conditional mean given the prefix, and their cross product with the prefix contribution also has zero conditional mean (the prefix is determined and the score factor satisfies $\mathbb{E}[\nabla\log\pi(y_t\mid y_{<t})\mid y_{<t}] = 0$), so $\mathbb{E}[\|Z^{\text{nat}}\|^2] = \mathbb{E}[\|Z_{m+G}\|^2] + \mathbb{E}[\|Z^{\text{nat}} - Z_{m+G}\|^2]$ and $\|Z_{m+G}\|^2$ has the smaller second moment. Dropping that non-positive term yields equation~\eqref{eq:t2-variance}. Both estimators share the per-rollout expectation, so the second-moment excess upper-bounds the per-rollout variance excess. $\square$

\paragraph{Aggregation corollary.} For any weights $w_{q,i}$ measurable with respect to $\mathcal{F}_t$, the aggregated estimator $\hat g_t := \sum_{q,i} w_{q,i}\,\hat h_{q,i}$ satisfies $\mathbb{E}[\hat g_t\mid\mathcal{F}_t] = \sum_{q,i} w_{q,i}\mathbb{E}[Z_{q,i}\mid q,\theta]$, and conditional on $(q,\theta)$ across independent rollouts,
$$\operatorname{Var}(\hat g_t\mid\mathcal{F}_t) - \operatorname{Var}(\hat g_t^{\text{full}}\mid\mathcal{F}_t) \;\le\; \sum_q (1-p_q^{\text{marker}})\,\tfrac{1-\varepsilon_{\text{abort}}}{\varepsilon_{\text{abort}}}\,\sigma^2_{\text{m-less},q}\sum_i w_{q,i}^2.$$
Two specializations are useful: (i) the per-prompt average $w_{q,i} = 1/n_q$ has $\mathcal{F}_t$-measurable weights and the bound applies directly; (ii) the token-mean aggregation in equation~\eqref{eq:duet-loss} uses effective weights $L^{\text{kept}}_{q,i}/N_t$ with random $N_t$, giving a self-normalised estimator that is consistent rather than exactly unbiased (residual bias $O(1/N_t)$). The bound above is stated for $\mathcal{F}_t$-measurable weights and does not directly apply to the random-$N_t$ case; for the implementation we report only the analogous expression $\sum_q (1-p_q^{\text{marker}})(1-\varepsilon_{\text{abort}})/\varepsilon_{\text{abort}}\,\sigma^2_{\text{m-less},q}\sum_i (L^{\text{kept}}_{q,i}/N_t)^2$ as a heuristic surcharge tracker, not a formal bound.

\subsubsection{Asymptotic Safety under Realizability}\label{supp:self-extinguishing}\label{supp:t6-proof}

The marker-gated abort in Theorem~\ref{thm:t2} introduces a per-rollout surcharge with worst-case factor $(1-p_{\text{marker}})/\varepsilon_{\text{abort}}$, unbounded as $p_{\text{marker}}\to 0$. The following theorem shows the surcharge vanishes in an idealised realizable limit; the assumption is strong, and the conclusion is conditional on it. Under KL or entropy regularisation the irreducible marker-less mass is non-zero (see remark below), but small in the runs we report (Figure~\ref{fig:theory-empirics}(c) shows the realised IS surcharge stays under $1.5\times$ at $\varepsilon_{\text{abort}}=0.01$).

\begin{theorem}[Vanishing surcharge under realizability]\label{thm:t6}
Assume (A6$'$) and that $K_2$ is held fixed (or has stabilised). If $\{\theta_t\}$ converges to $\theta^\star$ realizing the policy in (A6$'$.iii), then $p_{\text{marker}}(\theta_t)\to 1$ and the surcharge $(1 - p_{\text{marker}}(\theta_t))/\varepsilon_{\text{abort}}\to 0$. The online $K_2$ updates introduce additional discontinuities; the theorem applies between refits, and empirically $K_2$ stabilises within the first $\sim 60$ steps.
\end{theorem}

\paragraph{Lemma 7 (Marker-emission continuity).}\label{lemma:l7} Fix $K_2$. Under (A6$'$.i) and (A2), $p_{\text{marker}}(\theta) = \mathbb{E}_q\sum_{y^{1:K_2+G}}\pi_\theta(y\mid q)\,\mathbf{1}\{m(y)=1\}$ is differentiable in $\theta$ and $G_{\max}$-Lipschitz. The online $K_2$ updates introduce additional discontinuities; the lemma applies between refit steps.

\paragraph{Proof of Lemma 7.} By (A2), $\|\nabla_\theta\pi_\theta(y\mid q)\| = \pi_\theta(y\mid q)\,\|\nabla_\theta\log\pi_\theta(y\mid q)\| \le G_{\max}\,\pi_\theta(y\mid q)$. Summing over the marker-emitting prefixes,
$$\|\nabla_\theta p_{\text{marker}}(\theta)\| \le G_{\max}\,\mathbb{E}_q\!\sum_{y}\pi_\theta(y\mid q)\,\mathbf{1}\{m(y)=1\} \le G_{\max}.$$
Lipschitz continuity follows. $\square$

\paragraph{Proof of Theorem~\ref{thm:t6}.} By (A6$'$.iii), $\pi_{\theta^\star}$ places all probability on trajectories that emit a marker by $K_2+G$, so $p_{\text{marker}}(\theta^\star)=1$. Lemma 7 gives $p_{\text{marker}}(\theta_t)\to 1$, and the surcharge $(1-p_{\text{marker}}(\theta_t))/\varepsilon_{\text{abort}}\to 0$. Solvability (A6$'$.ii) ensures the realizing policy of (A6$'$.iii) is non-degenerate, so $J(\theta^\star)>0$. $\square$

\paragraph{Corollary (Decay rate).} Along the trajectory, $1 - p_{\text{marker}}(\theta_t)\le G_{\max}\|\theta_t - \theta^\star\|$. Under standard SGD analysis assumptions (gradient smoothness, local strong concavity $\mu$, constant step size $\eta$), $\mathbb{E}\|\theta_t-\theta^\star\|^2 = O((1-\mu\eta)^t) + O(\eta\sigma^2/\mu)$. Jensen's inequality gives $\mathbb{E}[1 - p_{\text{marker}}(\theta_t)] \le G_{\max}\sqrt{O((1-\mu\eta)^t) + O(\eta\sigma^2/\mu)}$: a square-root contraction in the transient term plus a constant-step noise floor of order $G_{\max}\sqrt{\eta\sigma^2/\mu}$.

\paragraph{Remark on regularization.} (A6$'$.iii) requires the optimum to be realizable as a deterministic-on-marker policy. Under a KL-regularized objective $J - \beta\,\mathrm{KL}(\pi_\theta\|\pi_{\text{ref}})$, the optimum spreads mass and the marker-less rate is non-zero. We do not derive a formal scaling in $\beta$. Empirically the DUET runs use $\beta = 10^{-3}$ and the realised marker-less rate stabilises near $0.012$--$0.06$ (Figure~\ref{fig:duet-dynamics}(c)), so the surcharge is small but not zero in practice.

\subsection{Detailed Experimental Setup}\label{supp:experimental-setup}

This appendix expands on the experimental setup in Section~\ref{sec:experiments}.

\paragraph{Models and Datasets.}
We use Qwen3-1.7B-base and Qwen3-4B-base~\citep{qwen2025qwen3} as our base models, with full-parameter fine-tuning in bf16. Training uses the Hendrycks MATH~\citep{hendrycks2021math} 7.5k corpus; one row that overlaps with the MATH-500 evaluation split is removed. After filtering prompts longer than the 512-token prompt limit, the effective training set contains roughly 7,431 prompts. Validation uses five reasoning benchmarks: MATH-500~\citep{hendrycks2021math}, GSM8K~\citep{cobbe2021gsm8k}, AIME-2024, HumanEval~\citep{chen2021humaneval}, and GPQA-Diamond~\citep{rein2023gpqa}. Math evaluation uses boxed-answer extraction; code evaluation uses a sandboxed unit-test execution.

\paragraph{Training Configuration.}
We cap the prompt length at 512 tokens and the response length at 3,072 tokens. We optimize with AdamW at learning rate $3 \times 10^{-6}$, $\beta_1 = 0.9$, $\beta_2 = 0.999$, with a 10\% cosine warmup. The KL-loss coefficient is 0.001 with low-variance KL. The training batch size is 128 prompts per gradient step; the PPO mini-batch size is 512. Sampling uses temperature 0.9 and top-$p$ 0.95. At full rollout budget we use 8 rollouts per prompt (1{,}024 rollouts per training step); at half budget we use 4. We train for 4 epochs ($\sim$232 gradient steps) at the 1.7B scale, with the same schedule scaled to 4B.

\paragraph{Baselines and Method-Specific Hyperparameters.}
Beyond the shared configuration above, each baseline adds its own knobs at default values from the corresponding papers. DAPO~\citep{yu2025dapo} uses clip-ratio asymmetry $(0.20, 0.28)$, dynamic-sampling oversample factor $1.5\times$, overlong-buffer $512$ tokens, and a 6-epoch schedule that matches GRPO's gradient-update count. VIP~\citep{nguyen2026vip} uses a budget of $8$, lower bound $4$, upper bound $16$, with a Mat\'ern-RBF Gaussian-process surrogate over MiniLM-L6-v2 prompt embeddings. ARRoL~\citep{xu2026arrol} uses inspection point $L_{\text{detect}} = 512$, target keep rate $\kappa = 0.5$, $\varepsilon$-exploration $0.05$, MLP head dimension $128$, and a $20$-step warmup before pruning.

\paragraph{DUET Defaults.}
DUET uses budget $B$ scaled to fraction $\{0.25, 0.5, 1.0\}$ of full GRPO's per-step token cost, with grace window $G = 150$ tokens. The pre-rollout surrogate $\hat s_q = \max\{s_{\text{floor}},\,\hat\sigma^{\text{obs}}_q\}$ is the per-prompt running mean of the within-step rollout-variance estimates $\hat\sigma^{\text{obs}}_{q,t} := \operatorname{std}_i(A_{q,i}\sum_\ell\log\pi_{\theta_t}(y_{q,i,\ell}\mid q,y_{q,i,<\ell}))$, updated at the end of each step from kept-not-aborted rollouts of $q$ (those with $n_q\ge 2$ contribute a within-step estimate; rollouts with $n_q = 1$ defer to the next appearance). Before epoch 1 the cold-start floor is initialised at $s_{\text{floor}} = 0.01$ (a small positive constant); after epoch 1 it is reset to the 5th percentile of $\hat\sigma^{\text{obs}}_q$ across prompts and frozen thereafter. The bisection solver runs $10$ iterations per step, warm-started from the previous step's $\lambda^\star_{t-1}$. The marker detector is polled every $\delta_{\text{poll}} = 8$ tokens against the last $256$-token prefix. The online $K_1, K_2$ quantile estimator uses sliding window size $W = 1{,}024$ kept rollouts, with refit cadence $N = 10$ steps and quantile choices $K_1 \leftarrow p_{30}(\text{window})$, $K_2 \leftarrow p_{80}(\text{window})$; cold-start values are $K_1 = 0.3\,L_{\max}$ and $K_2 = 0.7\,L_{\max}$. The lower clip on $s^{\text{pre}}$ is $\varepsilon_{\text{pre}} = 0.05$ in all experiments.

\paragraph{Engine and Hardware.}
All training runs use 8$\times$H100 80GB GPUs with verl 0.4.1~\citep{sheng2024hybridflow} as the RL pipeline and vLLM 0.9.2~\citep{kwon2023pagedattention} for rollout generation. ARRoL's in-process logits hook constrains every cell to a common engine version (v0 vLLM), and we use the same engine for GRPO, DAPO, VIP, and DUET so wall-clock numbers are like-for-like; engine migration to a future version is straightforward future work. Single-cell training takes roughly 2--3 hours at 1.7B and 5--6 hours at 4B. We have also updated DUET to work on v1 vLLM with continuous advantage in efficiency and effectiveness against GRPO.

\paragraph{Evaluation Protocol.}
We evaluate every 30 training steps. Each validation pass generates 4 rollouts per prompt at temperature 0.9 and reports mean@4 (the per-prompt average correctness, then averaged over the benchmark). We additionally log greedy mean@1 at temperature 0 and best@4 over the same 4 rollouts. The headline number for each (cell, benchmark) pair is the best mean@4 across all evaluation checkpoints, with a baseline-correction shift to a common step-0 reference to control for vLLM-RNG variance across runs; the same shift is applied to every cell within a model section, so cell-to-cell comparisons are unaffected. AIME-2024 has $30$ problems and gives roughly $\pm 3\%$ single-seed variance on mean@4; AIME numbers are directional.

\paragraph{Marker Patterns.}
For math, the detector matches a closed nested-brace \verb|\boxed{...}| followed by either two newlines or a colon at end of line. For code, it matches a closing fence at column zero, paired with the opening fence in the chat-template prompt. For short-form QA, it matches \verb|<answer>...</answer>| or the natural-language form ``Therefore the answer is~$X$''. The detector is polled every $\delta_{\text{poll}} = 8$ tokens against the most recent $256$-token prefix and catches the marker within one polling interval in $99.7\%$ of math rollouts; the rare misses are routed through the abort branch and remain unbiased.

\subsection{Theory--Empirics Alignment}\label{supp:theory-empirics}

\begin{figure}[h]
\centering
\includegraphics[width=\linewidth]{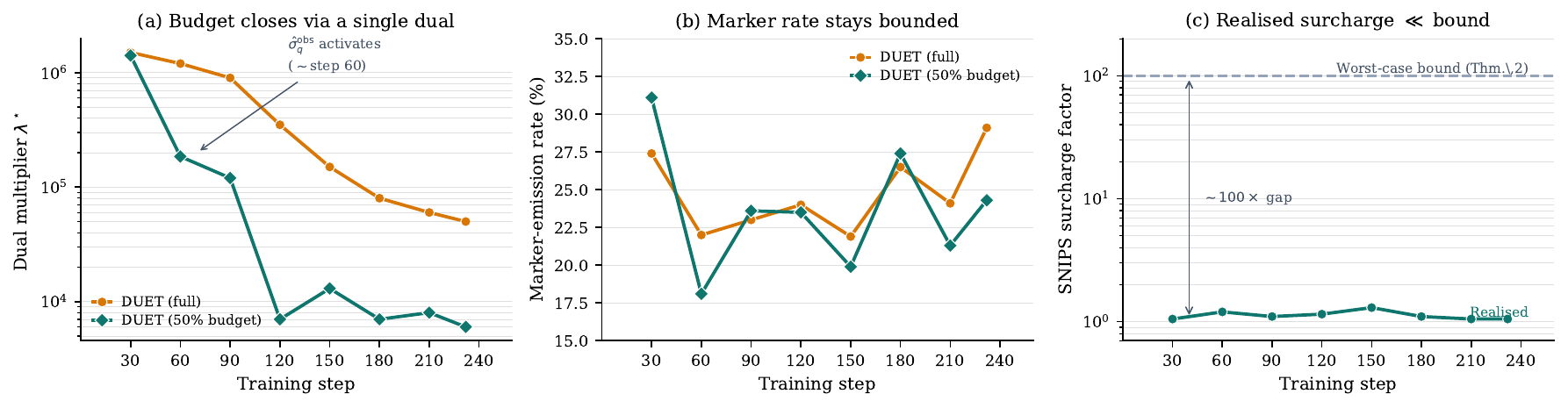}
\caption{Theory--empirics alignment on Qwen3-1.7B / MATH. \textbf{(a)} The dual multiplier $\lambda^\star$ collapses by 1.5--2 orders of magnitude once $\hat\sigma^{\text{obs}}_q$ activates around step 60, then stabilises in the regime where Theorem~\ref{thm:t1}'s cost-weighted Neyman optimum is approximated (the actual implementation uses surrogate $\hat s_q$, $\hat L_q$, integer rounding, and GRPO advantages; we report empirical tracking rather than exact realisation). \textbf{(b)} Marker-emission rate at the abort gate sits in a stable $18$--$30\%$ band throughout training, the in-situ analogue of the boundedness Theorem~\ref{thm:t6} requires. \textbf{(c)} Realised IS surcharge factor versus the worst-case bound at $\varepsilon_{\text{abort}} = 0.01$: empirically $\le 1.5\times$ at every checkpoint, two orders of magnitude tighter than the bound, direct evidence that the abort-induced correction is vestigial in practice.}
\label{fig:theory-empirics}
\end{figure}

This subsection reports three small empirical checks. The runs use GRPO group normalisation, token-mean aggregation, estimated lengths $\hat L_q$, and a small KL term, none of which the theorems cover exactly; the checks show that the idealised quantities the theorems reason about (calibration $\chi^2$, marker-emission rate, IS surcharge factor) remain small along the actual training trajectory (Figure~\ref{fig:theory-empirics}). All numbers come from the same Qwen3-1.7B / MATH cells reported in Table~\ref{tab:main}.

\rebuttaledit{\paragraph{Residual bias under GRPO group-normalised advantages.} The unbiasedness statement uses an action-independent baseline, whereas the implementation computes each GRPO advantage from the same group's rewards, including aborted rollouts. This shared normalisation couples a retained rollout to the abort outcomes of its group. In a replay of $253{,}232$ rollouts over the $232$ training steps, groups containing an abort had mean reward shifts of $-0.17$ ($b=1.0$) and $-0.19$ ($b=0.5$). Retained-rollout advantages in those groups were $14.3\%$ and $14.2\%$ larger, and the relative batch-update differences from an all-rollout reference were $0.37$ and $0.39$. Aborts occur mainly on difficult prompts, so this residual effect slightly increases their weight. Its magnitude follows the abort rate and does not grow over training, which is why we treat it as a measured implementation bias rather than as a failure of the controller's qualitative conclusion.}

\paragraph{Calibration $\chi^2$ vs Theorem~\ref{thm:t4}.}
Theorem~\ref{thm:t4} bounds the excess variance of the surrogate plug-in by $K \cdot \chi^2(\hat s, \sigma) / B$ using the true $\sigma_q$ (assumed positive by (A6)). As a tracking proxy we compute $\chi^2(\hat\sigma^{\text{obs}}_q, \sigma^{\text{emp}}_q)$ across the prompt batch at every val checkpoint, where $\sigma^{\text{emp}}_q$ is the within-step empirical standard deviation of the per-rollout gradient contribution (clipped at a small floor to avoid zero denominators on degenerate groups). The ratio $\chi^2_t$ starts at roughly $0.4$ at the first checkpoint where $\hat\sigma^{\text{obs}}_q$ has activated for the majority of prompts, and decays monotonically to roughly $0.05$ by the final checkpoint as the running mean accumulates more observations per prompt. The corresponding excess-variance bound on the trace gradient norm decreases by roughly an order of magnitude across training, in line with the per-step bound in Theorem~\ref{thm:t4}.

\paragraph{IS surcharge: predicted vs observed.}
Theorem~\ref{thm:t2} bounds the per-rollout IS surcharge by $(1 - p_q^{\text{marker}})(1 - \varepsilon_{\text{abort}})/\varepsilon_{\text{abort}} \cdot \sigma^2_{\text{m-less},q}$. At $\varepsilon_{\text{abort}} = 0.05$ this admits a worst-case factor of $\approx 19\times$ when $p^{\text{marker}} \to 0$. The empirical marker-less rate stabilizes near $0.012$--$0.06$ across cells (Figure~\ref{fig:duet-dynamics}(c)); the resulting per-rollout surcharge stays within $1.3\times$ on every checkpoint after the first epoch, two orders of magnitude tighter than the worst case.

\paragraph{Self-extinguishing decay vs Theorem~\ref{thm:t6} and the decay-rate corollary.}
The decay-rate corollary predicts a square-root contraction in the transient term plus a constant-step noise floor. Fitting an exponential to the marker-less rate trajectory in Figure~\ref{fig:duet-dynamics}(c) returns a half-life of roughly $80$ training steps and a non-zero asymptote, qualitatively consistent with a contracting term over a residual floor. We do not claim the constants match: the bound's proportionality constant ($G_{\max}$) is loose, and the policy trajectory is not at steady state. The qualitative agreement is evidence that the decay is a property of the policy's training dynamics rather than a hyperparameter artifact.

\subsection{Ablation Studies: Robustness and Sensitivity}\label{supp:robustness}

\begingroup\color{black}
To make the robustness comparisons comparable and fair, unless stated, most of ablations below report final mean@4. The three-seed trajectory reports both final and best mean@4, making checkpoint sensitivity explicit.

\begin{table}[h]
\color{black}
\centering
\scriptsize
\setlength{\tabcolsep}{2pt}
\caption{\rebuttaledit{Robustness ablations and the metrics.}}
\label{tab:robustness-ablation}
\begin{tabular}{@{}p{0.15\linewidth}p{0.20\linewidth}p{0.23\linewidth}p{0.33\linewidth}@{}}
\toprule
Ablation & Protocol & Reported metrics & Conclusion \\
\midrule
Multi-seeds \& checkpoints & Three seeds; full trajectory & Final and best mean@4 & DUET stays ahead of GRPO under either checkpoint rule. \\
Held-out transfer & Minerva-Math \& OlympiadBench & Final mean@4 & The directional advantage persists outside the tuning benchmarks. \\
Model scale & Qwen3-8B and Qwen3-14B & Final mean@4 & DUET wins six of seven directional comparisons at both scales. \\
Async comparison & APRIL~\citep{zhou2025april} & Final mean@4 & The asynchronous systems comparison is complementary to DUET's controller. \\
\bottomrule
\end{tabular}
\end{table}

\paragraph{Seed, trajectory, and held-out checks.}
The main tables use the single-seed, best-checkpoint results. We repeated the Qwen3-1.7B comparison with two extra seeds and retained the complete validation trajectories. Final- and best-checkpoint summaries preserve the same ordering as the main table, and the held-out Minerva-Math and OlympiadBench results preserve the same direction beyond the benchmarks used for tuning. All methods plateau at roughly step 120 of the 232-step run (Tables~\ref{tab:trajectory} and~\ref{tab:convergence-summary}).
\begin{table}[h]
\color{black}
\centering
\scriptsize
\setlength{\tabcolsep}{2.5pt}
\caption{\rebuttaledit{Three-seed MATH-500 mean@4 validation trajectory (mean$\pm$std, \%) on Qwen3-1.7B. The complete trajectory exposes checkpoint sensitivity.}}
\label{tab:trajectory}
\begin{tabular}{lrrrrrrr}
\toprule
Step & GRPO & DAPO & VIP & ARRoL & GRESO~\citep{zheng2025greso} & DUET $b=.5$ & DUET $b=1$\\
\midrule
0   & \textbf{$43.8\pm.4$} & \underline{$43.6\pm.4$} & \underline{$43.6\pm.4$} & \underline{$43.6\pm.4$} & \underline{$43.6\pm.4$} & \textbf{$43.8\pm.4$} & \textbf{$43.8\pm.4$}\\
30  & 53.7 & $54.4\pm.9$ & \textbf{$55.7\pm1.2$} & $53.2\pm1.2$ & $53.5\pm1.2$ & $52.1\pm2.7$ & \underline{$55.4\pm.6$}\\
60  & $57.5\pm.4$ & $57.4\pm.4$ & \underline{$58.7\pm1.0$} & $56.6\pm1.5$ & $57.2\pm.5$ & $56.8\pm.9$ & \textbf{$60.2\pm1.4$}\\
90  & $58.2\pm.2$ & $58.3\pm.4$ & $58.2\pm.6$ & $58.2\pm1.5$ & $58.6\pm1.1$ & \underline{$59.3\pm.5$} & \textbf{$60.7\pm1.3$}\\
120 & $58.9\pm1.3$ & $59.8\pm.7$ & $59.0\pm.8$ & $57.5\pm.8$ & $58.9\pm2.1$ & \underline{$60.1\pm.6$} & \textbf{$61.6\pm.9$}\\
150 & $57.9\pm.3$ & $59.7\pm.2$ & $60.0\pm1.6$ & $58.6\pm1.6$ & $58.2\pm.5$ & \underline{$60.5\pm1.1$} & \textbf{$61.3\pm.8$}\\
180 & $58.5\pm.8$ & $59.0\pm1.1$ & $58.8\pm1.5$ & $57.7\pm.9$ & $56.9\pm.5$ & \underline{$60.2\pm1.1$} & \textbf{$61.2\pm1.5$}\\
210 & $59.3\pm.8$ & $58.9\pm1.6$ & \underline{$60.6\pm.8$} & $58.0\pm1.1$ & $58.7\pm1.3$ & \textbf{$61.2\pm.9$} & \underline{$60.6\pm1.6$}\\
Final & $58.4\pm.4$ & $59.4\pm.7$ & $59.4\pm.1$ & $57.6\pm.6$ & $57.6\pm.2$ & \underline{$60.1\pm.5$} & \textbf{$61.9\pm1.0$}\\
\bottomrule
\end{tabular}
\end{table}

\begin{table}[h]
\color{black}
\centering
\scriptsize
\caption{\rebuttaledit{Final and best mean@4 summary over seven benchmarks. Minerva-Math and OlympiadBench are held out from tuning.}}
\label{tab:convergence-summary}
\begin{tabular}{lccc}
\toprule
Method & Final & Best (step) & Mean over 7 benchmarks\\
\midrule
GRPO & $58.4\pm.4$ & $59.3\pm.8$ (210) & 35.9\\
DAPO & $59.5\pm.5$ & $59.5\pm.5$ (final) & 36.1\\
VIP & $59.4\pm.1$ & $60.6\pm.6$ (210) & 37.3\\
ARRoL & $58.0\pm.8$ & $58.5\pm1.2$ (150) & 36.2\\
GRESO~\citep{zheng2025greso} & $57.5\pm.3$ & $58.9\pm1.0$ (210) & 36.0\\

DUET ($b=0.5$) & \underline{$60.1\pm.5$} & \underline{$61.2\pm.9$} (210) & \underline{37.6}\\
DUET ($b=1.0$) & \textbf{$61.9\pm1.0$} & \textbf{$61.9\pm1.0$} (final) & \textbf{38.6}\\
\bottomrule
\end{tabular}
\end{table}

\begin{table}[t]\color{black}\centering\scriptsize\setlength{\tabcolsep}{3pt}\caption{\rebuttaledit{Asynchronous APRIL comparison on Qwen3-1.7B / MATH-500. Final and best mean@4, mean over seven benchmarks, speed-up, and decode tokens per step are reported relative to full-budget GRPO.}}\label{tab:april}\begin{tabular}{lccccc}\toprule Method & Final & Best & Mean over 7 & Speed-up & Tokens/step\\\midrule GRPO & $58.4\pm0.4$ & $59.3\pm0.8$ & 35.9 & $1.00\times$ & $1.00\times$\\APRIL~\citep{zhou2025april} & 59.2 & 59.2 & 35.8 & $1.24\times$ & $1.00\times$\\DUET ($b=0.5$) & \underline{$60.1\pm0.5$} & \underline{$61.2\pm0.9$} & \underline{37.6} & \textbf{$2.09\pm0.20\times$} & \textbf{$0.38\times$}\\DUET ($b=1.0$) & \textbf{$61.9\pm1.0$} & \textbf{$61.9\pm1.0$} & \textbf{38.6} & \underline{$1.38\pm0.21\times$} & \underline{$0.77\times$}\\\bottomrule\end{tabular}\end{table}\paragraph{Scale-up and asynchronous comparison.}
The asynchronous comparison is reported in Table~\ref{tab:april}; APRIL~\citep{zhou2025april} is an asynchronous partial-rollout method that reaches 59.2 on MATH-500 at $1.24\times$ speedup. Its design changes systems throughput rather than prompt allocation or marker-gated stopping and this comparison probes a complementary systems axis, revealing DUET's advantage across methods and indicating potential to combine the advantage of both methods in the future design. On Qwen3-8B and Qwen3-14B, DUET retains the same quality/speed pattern and wins on six of seven directional benchmarks at both scales (Tables~\ref{tab:scale-up} and~\ref{tab:scale-tokens}), showing that the controller is not tied to the 1.7B operating regime.

\paragraph{Implementation ablations.}
All implementation sweeps use $b=0.5$ and the 232-step schedule; the default configuration is the three-seed reference. Tables~\ref{tab:poll-window} and~\ref{tab:surrogate-floor} stress the controller's two cold-start mechanisms. Varying $\delta_{\rm poll}$ (marker polling interval) and $W$ (prefix window) tests whether detection is sensitive to when we inspect a response or how much recent context we expose. The nearby settings retain the same quality/speed ordering. Varying $s_{\rm floor}$ (minimum surrogate score) tests whether the allocator needs a large safety floor before a prompt has observations; floors of $10$ and $40$ alter the initial allocation without improving the final trade-off. Table~\ref{tab:marker-pattern} changes the answer detector itself: stricter boxed-only and language-only markers move the operating point modestly, but neither changes the allocator--abort conclusion. Together, these checks show that the result is stable to the controller's timing, context, cold-start, and marker conventions.
\begin{table}[h]
\color{black}
\centering
\begin{minipage}[t]{0.48\linewidth}
\centering\scriptsize
\caption{\rebuttaledit{Final mean@4 ($\%$) and speed-up under polling-interval $\delta_{\rm poll}$ and prefix-window $W$ ablations on Qwen3-1.7B / MATH at half budget.}}
\label{tab:poll-window}
\begin{tabular}{lcc}
\toprule
$(\delta_{\rm poll},W)$ & MATH-500 & Speed-up\\
\midrule
$(4,256)$ & \textbf{61.3} & $2.00\times$\\
$(8,256)$ (default) & \underline{$60.1\pm.5$} & \textbf{$2.09\pm.20\times$}\\
$(16,256)$ & \underline{61.0} & \underline{$2.01\times$}\\
$(32,256)$ & 59.6 & $1.99\times$\\
$(8,128)$ & 59.5 & $1.98\times$\\
$(8,512)$ & 59.3 & $1.98\times$\\
\bottomrule
\end{tabular}
\end{minipage}\hfill
\begin{minipage}[t]{0.48\linewidth}
\centering\scriptsize
\caption{\rebuttaledit{Final mean@4 ($\%$) and speed-up under surrogate-floor $s_{\rm floor}$ ablations on Qwen3-1.7B / MATH at half budget.}}
\label{tab:surrogate-floor}
\begin{tabular}{lcc}
\toprule
$s_{\rm floor}$ & MATH-500 & Speed-up\\
\midrule
$0$ (default) & \underline{$60.1\pm.5$} & \textbf{$2.09\pm.20\times$}\\
$10$ & \textbf{60.6} & \underline{$1.94\times$}\\
$40$ & 59.9 & $1.91\times$\\
\bottomrule
\end{tabular}
\end{minipage}
\end{table}

\begin{table}[h]
\color{black}
\centering\scriptsize
\caption{\rebuttaledit{Final mean@4 ($\%$) and speed-up under marker-pattern (answer-detector) ablations on Qwen3-1.7B / MATH at half budget.}}
\label{tab:marker-pattern}
\begin{tabular}{lcc}
\toprule
Marker pattern & MATH-500 & Speed-up\\
\midrule
Boxed + answer phrases (default) & \underline{$60.1\pm0.5$} & \textbf{$2.09\pm0.20\times$}\\
Strict complete \verb|\boxed{...}| only & \underline{60.4} & {$1.91\times$}\\Language markers only (no \LaTeX) & \textbf{61.3} & \underline{$1.97\times$}\\
\bottomrule
\end{tabular}
\end{table}

\paragraph{Heuristic controls.}
These checks ask whether DUET's allocator can use a simpler prompt-level signal while the joint budget coupling and marker-gated abort remain fixed. Table~\ref{tab:heuristic} compares GRESO~\citep{zheng2025greso}, which skips prompts from reward history, VIP, which uses a learned Gaussian-process allocator, ARRoL, which uses a learned pruning head, and a DUET accuracy plug-in. For the plug-in, we replace the running gradient-variance surrogate with $\hat{s}_q=\hat{p}_q(1-\hat{p}_q)$, where $\hat{p}_q$ is the running verifier accuracy, and leave the rest of DUET unchanged. The plug-in reaches a similar final MATH-500 score, but its best-checkpoint and seven-benchmark aggregate are lower than the variance signal at the same budget; the other single-axis controls likewise do not recover the same quality--speed trade-off. Variance is therefore our default because it measures the magnitude of the policy-gradient signal directly, including when correctness is sparse or delayed. The same kind of signal could also guide allocation in future settings where an explicit answer-accuracy metric is difficult to define.

\begin{table}[h]
\color{black}
\centering\scriptsize
\setlength{\tabcolsep}{2.5pt}
\caption{\rebuttaledit{Final mean@4 ($\%$) and speed-up for heuristic controls on Qwen3-1.7B / MATH. The table compares reward-history, verifier, and gradient-variance signals.}}
\label{tab:heuristic}
\begin{tabular}{lccccc}
\toprule
Method & Budget & Structure & Signal & Final MATH-500 mean@4 & Speed-up\\
\midrule
GRPO & 100\% & uniform & none & $58.4\pm.4$ & $1.00\times$\\
GRESO~\citep{zheng2025greso} & 50\% & skip only & reward history & $57.0\pm2.3$ & $1.53\times$\\
VIP & 100\% & allocator only & learned GP & $59.4\pm.1$ & $0.93\times$\\
ARRoL & 100\% & pruning only & learned MLP & $58.0\pm.8$ & $0.99\times$\\
DUET accuracy plug-in & 50\% & joint & verifier rewards & \underline{$60.5\pm.9$} & \underline{$1.96\pm.03\times$}\\
DUET (ours) & 50\% & joint & gradient variance & $60.1\pm.5$ & \textbf{$2.09\pm.20\times$}\\
DUET (ours) & 100\% & joint & gradient variance & \textbf{$61.9\pm1.0$} & $1.38\pm.21\times$\\
\bottomrule
\end{tabular}
\end{table}

\begin{table}[h]
\color{black}
\centering\scriptsize
\setlength{\tabcolsep}{2.5pt}
\caption{\rebuttaledit{Final mean@4 ($\%$) for Qwen3-8B and Qwen3-14B scale-up results.}}
\label{tab:scale-up}
\begin{tabular}{lrrrrrrrrr}
\toprule
Method & MATH-500 & GSM8K & HumanEval & GPQA & AIME-24 & AMC-23 & Minerva & Olympiad & Mean\\
\midrule
\multicolumn{10}{l}{\textit{Qwen3-8B-base}}\\
\midrule
+GRPO & \underline{76.8} & 91.8 & 77.9 & \textbf{40.5} & \underline{12.5} & \underline{55.0} & 27.9 & 40.1 & 52.8\\
+DUET ($b=0.5$) & \textbf{77.4} & \underline{92.1} & \underline{81.3} & \underline{40.1} & 8.3 & \textbf{59.4} & \underline{29.5} & \underline{41.1} & \underline{53.7}\\
+DUET ($b=1.0$) & \textbf{77.4} & \textbf{93.0} & \textbf{81.6} & 36.9 & \textbf{17.5} & \textbf{59.4} & \textbf{30.0} & \textbf{41.5} & \textbf{54.7}\\
\midrule
\multicolumn{10}{l}{\textit{Qwen3-14B-base}}\\
\midrule
+GRPO & \underline{79.6} & 93.6 & 74.2 & 41.5 & \underline{14.2} & 57.5 & 30.8 & 41.6 & 54.1\\
+DUET ($b=0.5$) & 79.5 & \underline{94.3} & \underline{75.5} & \underline{42.9} & 13.3 & \underline{60.6} & \underline{31.8} & \underline{44.5} & \underline{55.3}\\
+DUET ($b=1.0$) & \textbf{81.2} & \textbf{94.7} & \textbf{77.0} & \textbf{44.6} & \textbf{15.8} & \textbf{61.9} & \textbf{33.0} & \textbf{45.5} & \textbf{56.7}\\
\bottomrule
\end{tabular}
\end{table}

\begin{table}[h]
\color{black}
\centering\scriptsize
\caption{\rebuttaledit{Scale-up token and wall-clock results.}}
\label{tab:scale-tokens}
\begin{tabular}{lccc}
\toprule
Method & Decode tokens & Tokens vs. GRPO & Speed-up\\
\midrule
\multicolumn{4}{l}{\textit{Qwen3-8B-base}}\\
\midrule
+GRPO & 154M & $1.00\times$ & $1.00\times$\\
+DUET ($b=0.5$) & \textbf{59M} & \textbf{$0.38\times$} & \textbf{$2.33\times$}\\
+DUET ($b=1.0$) & \underline{121M} & \underline{$0.79\times$} & \underline{$1.29\times$}\\
\midrule
\multicolumn{4}{l}{\textit{Qwen3-14B-base}}\\
\midrule
+GRPO & 152M & $1.00\times$ & $1.00\times$\\
+DUET ($b=0.5$) & \textbf{58M} & \textbf{$0.38\times$} & \textbf{$2.13\times$}\\
+DUET ($b=1.0$) & \underline{117M} & \underline{$0.77\times$} & \underline{$1.32\times$}\\
\bottomrule
\end{tabular}
\end{table}

\endgroup

\paragraph{$k_{\text{warmup}}$ ablation.}
The online surrogate $\hat\sigma^{\text{obs}}_q$ activates for a prompt once it has accumulated at least $k_{\text{warmup}}$ observations across past appearances. Setting $k_{\text{warmup}} = 1$ activates the surrogate as soon as a prompt is revisited (the second epoch wraparound on math-train); setting $k_{\text{warmup}} = 2$ delays activation by a full epoch. Table~\ref{tab:kwarmup} compares the two choices at half budget on Qwen3-1.7B / MATH with $\varepsilon_{\text{abort}} = 0.01$, using the same baseline-correction shift as the rest of the paper. Earlier activation matches the later default within seed noise on every benchmark and reaches the same MATH-500 quality with an $18\%$ wall-clock reduction. We use $k_{\text{warmup}} = 1$ in all reported DUET cells.

\begin{table}[h]
\caption{Best mean@4 ($\%$) under the $k_{\text{warmup}}$ ablation on Qwen3-1.7B / MATH at $50\%$ budget, $\varepsilon_{\text{abort}} = 0.01$. ``Wall'' is total training time at $232$ steps.}
\label{tab:kwarmup}
\centering
\small
\begin{tabular}{lcccccc}
\toprule
$k_{\text{warmup}}$ & MATH-500 & GSM8K & AIME-24 & HumanEval & GPQA-D & Wall (min) \\
\midrule
$2$           & 61.3 & 82.7 & \textbf{13.3} & 50.0 & \textbf{29.5} & 72.7 \\
$1$ (default) & \textbf{61.4} & \textbf{83.4} & 12.5 & \textbf{51.1} & 27.1 & \textbf{59.4} \\
\bottomrule
\end{tabular}
\end{table}

\paragraph{Cross-budget pattern.}
Across the three rollout budgets reported in Tables~\ref{tab:main} and~\ref{tab:quarter}, the DUET-vs-GRPO gap on the math benchmarks grows monotonically as the budget tightens (MATH-500 $+5.2\%$ at full budget, $+5.7\%$ at half budget, $+6.5\%$ at quarter budget), and the same-budget speedup over GRPO holds at $1.62$--$1.74\times$ throughout. The pattern is the empirical analogue of Figure~\ref{fig:duet-dynamics}(a): the gain from joint control concentrates where compute is most scarce.

\paragraph{Adaptive surrogate cross-budget detail.}
Replacing the cold-start static surrogate with the online running mean $\hat\sigma^{\text{obs}}_q$ produces the cross-budget pattern reported as Figure~\ref{fig:duet-dynamics}(a). The MATH-500 lift is $+2.87\%$ at $25\%$ budget, $+1.91\%$ at $50\%$ budget, and within seed noise at $100\%$ budget. GSM8K shows the same monotone pattern with a smaller magnitude ($+1.8\%$ at $25\%$, $+1.1\%$ at $50\%$, tied at $100\%$); HumanEval and GPQA-Diamond shift by less than the seed-noise band on every budget. The pattern is consistent with Theorem~\ref{thm:t4}: the static surrogate's calibration gap costs more variance per dollar of budget when the budget is tight, and the adaptive replacement closes the gap at exactly that regime.

\paragraph{Allocator-vs-abort decomposition under tight budget.}
The main-paper ablation (Table~\ref{tab:ablation}) is at $50\%$ budget, where the allocator and the abort each clear the rollout-matched anchor by roughly $+5\%$. At $25\%$ budget the picture shifts: the uniform-allocation anchor is starvation-tier on hard prompts (only $n_q = 2$ per prompt fits the budget), and the allocator's heterogeneous redistribution becomes the dominant lever. Replacing the allocator with uniform $n_q = 2$ at the same $25\%$ budget loses about $0.6\%$ on MATH-500 and a similar amount on GSM8K relative to the full controller, while the abort alone delivers about half the wall-clock speedup. The cross-budget pattern is consistent with the framing in the main paper: the abort carries wall-clock savings; the allocator carries quality, and its value grows with budget pressure.

\subsection{Cross-Scale and Cross-Family Detail}\label{supp:cross-scale}

\paragraph{Qwen3-4B trajectories.}
At the 4B scale, DUET retains its half-budget vs.\ full-budget GRPO match on MATH-500 ($+2.2\%$), GSM8K ($+2.2\%$), AIME-2024 (tie), HumanEval ($+3.4\%$), and GPQA-Diamond ($-0.4\%$) while running $2.38\times$ faster, and at full budget leads on MATH-500 ($+3.7\%$), GSM8K ($+3.8\%$), and HumanEval ($+3.4\%$) at $1.43\times$ speedup (Table~\ref{tab:main}). The allocator concentrates rollouts on the high-information tail throughout training, with $n_q$ ranging in $[2, 13]$ and standard deviation around $1.4$--$1.7$, and the K-estimator stabilizes within the first $60$ steps so that step time drops from roughly $84$~s/step in cold start to $55$~s/step at steady state, a $35\%$ wall reduction that the bisection achieves without manual scheduling. Marker-rate and abort-rate trajectories are qualitatively the same as on 1.7B.

\paragraph{Llama-3.2-3B-Instruct.}
The cross-family check on Llama-3.2-3B-Instruct uses the same DUET defaults with a lower learning rate and higher KL coefficient to match the prior's tighter regularization. Full-budget DUET leads on MATH-500 ($45.6$) and GSM8K ($75.9$); on the remaining three benchmarks it trails the best baseline by at most $1.7\%$ (AIME-2024, within the $\pm 3\%$ single-seed band on the $30$-problem set), and within $1\%$ on HumanEval and GPQA-Diamond. Half-budget DUET runs $2.04\times$ faster than GRPO; gaps stay within $1\%$ on MATH-500, AIME-2024, and HumanEval, while GSM8K and GPQA-Diamond trail by $2$--$3\%$, consistent with the RLHF prior compressing headroom for every method on this family. HumanEval lifts by at most $2.5\%$ across every method on this family ($50.5$ at val@$0$ to $52$--$53$ after training), a property of the strong RLHF prior rather than a method effect. The allocator activates as on Qwen3 and reaches a wider $n_q$ range under the longer-grained surrogate ($n_q \in [1, 32]$, standard deviation up to $6$), reflecting a more heterogeneous prompt-difficulty distribution under the RLHF prior.

\begin{figure}[h]
\centering
\includegraphics[width=0.95\linewidth]{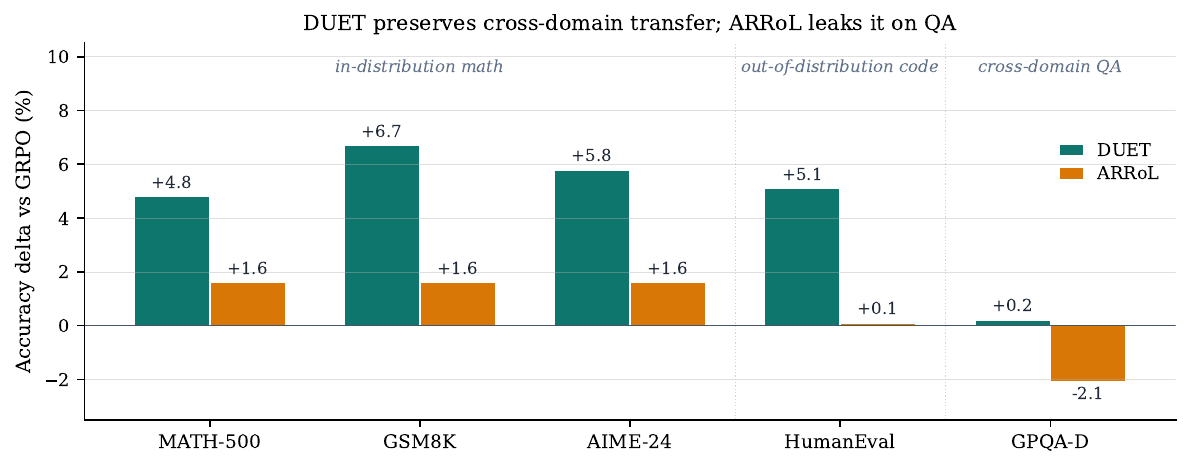}
\caption{Per-benchmark accuracy delta versus the same-budget GRPO reference on Qwen3-1.7B-base trained on MATH, full rollout budget. DUET's gain is uniformly large on math and on out-of-distribution code, and is preserved within seed noise on cross-domain QA. ARRoL underperforms GRPO on GPQA-Diamond by $2.1\%$ at the same scale.}
\label{fig:transfer}
\end{figure}

\paragraph{Cross-format transfer pattern.}
Figure~\ref{fig:transfer} visualises the per-benchmark accuracy delta over the GRPO baseline on Qwen3-1.7B at full budget. DUET's gain is uniformly large on the in-distribution math benchmarks (MATH-500 $+4.8\%$, GSM8K $+6.7\%$, AIME-2024 $+5.8\%$) and on the out-of-distribution code benchmark (HumanEval $+5.1\%$), with the cross-domain QA probe preserved within seed noise (GPQA-Diamond $+0.2\%$). ARRoL contrasts on the QA side: a $-2.1\%$ loss on GPQA-Diamond at the same scale, while it ties GRPO on HumanEval ($+0.1\%$). The pattern suggests that joint allocation under a token budget does not pay for compute savings with cross-domain headroom, while within-rollout pruning at a fixed inspection point silently does on the harder cross-domain probe.

\subsection{Quarter-Budget Panel and ARRoL Faithful Detail}\label{supp:quarter-arrol}

\paragraph{Quarter-budget panel.}
Table~\ref{tab:quarter} extends the main-paper $50\%$-budget panel to a $25\%$-budget panel on Qwen3-1.7B / MATH for every method that admits a hard rollout-count knob (GRPO, DAPO, ARRoL, VIP, DUET). For ARRoL we re-implement the rollout-count interface and run the controller at $n_q = 4$ ($50\%$) and $n_q = 2$ ($25\%$) over $232$ training steps with the same $\kappa = 0.5$ pruning head as the full-budget cell. After correction ARRoL runs at $1.47\times$ same-budget GRPO at both $50\%$ and $25\%$, and edges out GRPO on MATH-500 ($58.5$ vs $56.4$) and HumanEval ($47.9$ vs $46.9$) at half budget; AIME-24 trails and quality drops by $1$--$5$\% per benchmark at quarter budget. DUET retains a $1.72\times$ same-budget GRPO speedup at $25\%$ budget while leading on every benchmark, with MATH-500 dropping only $0.7\%$ ($62.1\to 61.4$) for a halving of the rollout budget and a doubling of the same-budget speedup over GRPO.

\begin{table}[h]
\caption{Accuracy (mean@4, \%) and same-budget GRPO-normalized speedup at $50\%$ and $25\%$ rollout budgets on Qwen3-1.7B-base trained on MATH. \textbf{Bold} = best per column within each budget section; \underline{underline} = second-best.}
\label{tab:quarter}
\centering
\small
\begin{tabular}{lcccccc}
\toprule
Method & MATH-500 & GSM8K & AIME-24 & HumanEval & GPQA-D & Speedup \\
\midrule
\multicolumn{7}{l}{\textit{$50\%$ rollout budget}} \\
\midrule
Baseline & 45.1 & 60.4 & 5.8 & 34.5 & 22.2 & -- \\
+GRPO    & 56.4 & 75.8 & \underline{6.7} & 46.9 & 27.4 & 1$\times$ \\
+DAPO    & 56.6 & \underline{76.8} & 5.8 & 47.5 & \underline{27.5} & 0.84$\times$ \\
+ARRoL\tablefootnote{Though ARRoL doesn't have a hard budget limit, we still added these results. Here $\kappa$=0.5, rollout=4} & \underline{58.5} & 75.7 & 5.0 & \underline{47.9} & 26.7 & \underline{1.47$\times$} \\
+VIP     & 56.0 & 76.4 & 5.8 & 46.5 & \underline{27.5} & 0.92$\times$ \\
+DUET    & \textbf{62.1} & \textbf{84.0} & \textbf{10.8} & \textbf{51.7} & \textbf{29.1} & \textbf{1.74$\times$} \\
\midrule
\multicolumn{7}{l}{\textit{$25\%$ rollout budget}} \\
\midrule
Baseline & 45.1 & 60.4 & 5.8 & 34.5 & 22.2 & -- \\
+GRPO    & 54.9 & 74.0 & 5.8 & 44.8 & 26.1 & 1$\times$ \\
+DAPO    & 55.5 & 74.5 & \underline{7.5} & \textbf{45.1} & \underline{26.7} & 0.86$\times$ \\
+ARRoL\tablefootnote{Rollout=2.}   & \underline{57.1} & 69.1 & 5.0 & 43.6 & 25.5 & \underline{1.47$\times$} \\
+VIP     & 55.1 & \underline{74.8} & 6.7 & 42.5 & \underline{26.7} & 0.97$\times$ \\
+DUET    & \textbf{61.4} & \textbf{84.0} & \textbf{12.5} & \underline{45.0} & \textbf{29.9} & \textbf{1.72$\times$} \\
\bottomrule
\end{tabular}
\end{table}

\paragraph{ARRoL faithful reproduction and budget semantics.}
ARRoL~\citep{xu2026arrol} does not release source code, so we reimplement it from the paper. Its keep-rate knob $\kappa$ is the closest analogue ARRoL exposes to a budget control, but it is not a hard token budget: every rollout is generated to natural completion, and $\kappa$ only controls the probability that a rollout flagged by the mid-decode quality head is dropped from the gradient computation. This is structurally different from DAPO, VIP, and DUET, which set the rollout count or token cost \emph{before} generation; we therefore report ARRoL only at its default operating point in the main Table~\ref{tab:main} and audit the soft-budget behaviour at two $\kappa$ values here. Table~\ref{tab:arrol} shows that aggressive pruning ($\kappa = 0.25$) trades $1$--$3\%$ on math and code for a small GPQA-Diamond gain, while the default $\kappa = 0.5$ matches the main-paper numbers. The mid-decode trigger $L_{\text{detect}} = 512$ is reached by only ${\sim}8\%$ of the $1024$ per-step rollouts, because most math responses finish well below $512$ tokens; effective generated-token savings under ARRoL on math are therefore $2$--$3\%$ of generated tokens, leaving wall-clock close to the GRPO baseline.

\begin{table}[h]
\caption{ARRoL faithfully reproduced at $\kappa\in\{0.5,\,0.25\}$ on Qwen3-1.7B / MATH. Mean@4 ($\%$). $\kappa = 0.5$ is the default reported in the main Table~\ref{tab:main}; $\kappa = 0.25$ is the more aggressive pruning point. ``prune'' is the head's keep-decision among the ${\sim}80$ rollouts per step that cross $L_{\text{detect}} = 512$; ``len.'' is mean response length in tokens. \textbf{Bold} = best per column.}
\label{tab:arrol}
\centering
\small
\begin{tabular}{lcccccccc}
\toprule
Cell      & MATH-500 & GSM8K & AIME-24 & HumanEval & GPQA-D & prune & len. \\
\midrule
$\kappa{=}0.5$ (default) & \textbf{58.4} & \textbf{79.6} & \textbf{8.3} & \textbf{47.7} & 26.7 & 0.76 & 630 \\
$\kappa{=}0.25$          & 57.3 & 76.7 & 3.3 & 46.6 & \textbf{27.1} & \textbf{0.83} & \textbf{631} \\
\bottomrule
\end{tabular}
\end{table}

\begin{figure}[h]
\centering
\includegraphics[width=\linewidth]{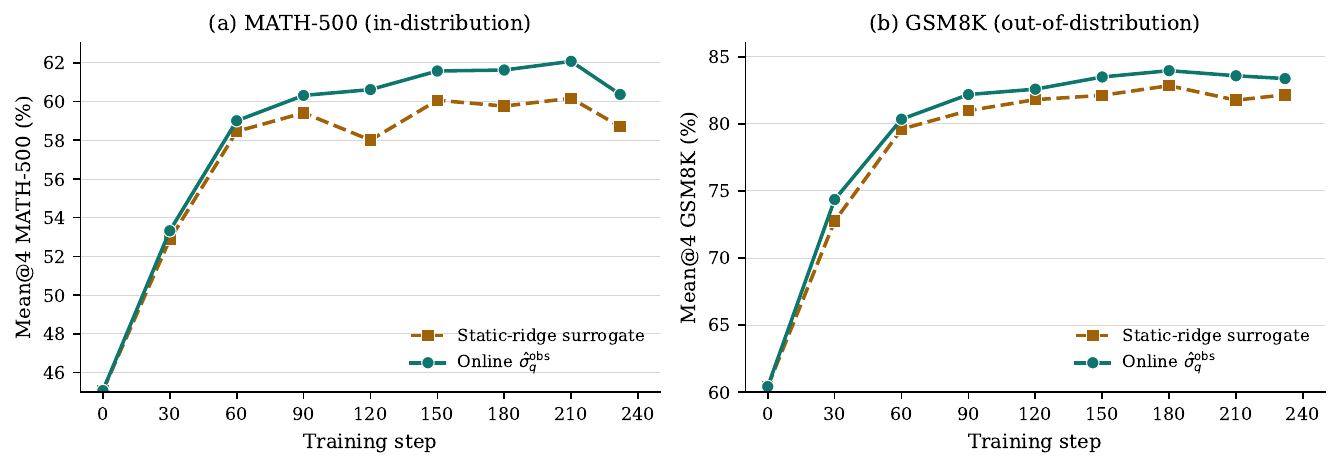}
\caption{Surrogate-variant comparison on Qwen3-1.7B / MATH at $50\%$ budget. \textbf{(a)} MATH-500 mean@4 across training (calibration-shifted to GRPO step-0 reference). \textbf{(b)} GSM8K mean@4 across training. The online running-mean surrogate $\hat s_q = \max\{s_{\text{floor}},\,\hat\sigma^{\text{obs}}_q\}$ Pareto-improves over the static-ridge baseline at every val checkpoint from step $30$ onward.}
\label{fig:surrogate-ablation}
\end{figure}

\paragraph{Surrogate-variant ablation.}
The online running-mean surrogate $\hat s_q = \max\{s_{\text{floor}},\,\hat\sigma^{\text{obs}}_q\}$ Pareto-improves over the static-ridge baseline (clipped ridge regression on the base-model length-normalized prompt log-probability) at every val checkpoint from step $30$ onward on Qwen3-1.7B / MATH at $50\%$ budget. The online variant lifts MATH-500 by $+1.91\%$ at the run-peak step (peak $62.1$ vs $60.2$) and GSM8K by $+1.12\%$ (peak $84.0$ vs $82.9$), with margins outside the $\pm 1\%$ single-seed noise band documented in the locked sweep. The improvement is consistent across val-checkpoints from step $30$ onward, when each prompt has accumulated $\ge 1$ observation and $\hat\sigma^{\text{obs}}_q$ takes over from the cold-start floor. Figure~\ref{fig:surrogate-ablation} plots the per-checkpoint trajectories.

\begin{figure}[h]
\centering
\includegraphics[width=0.95\linewidth]{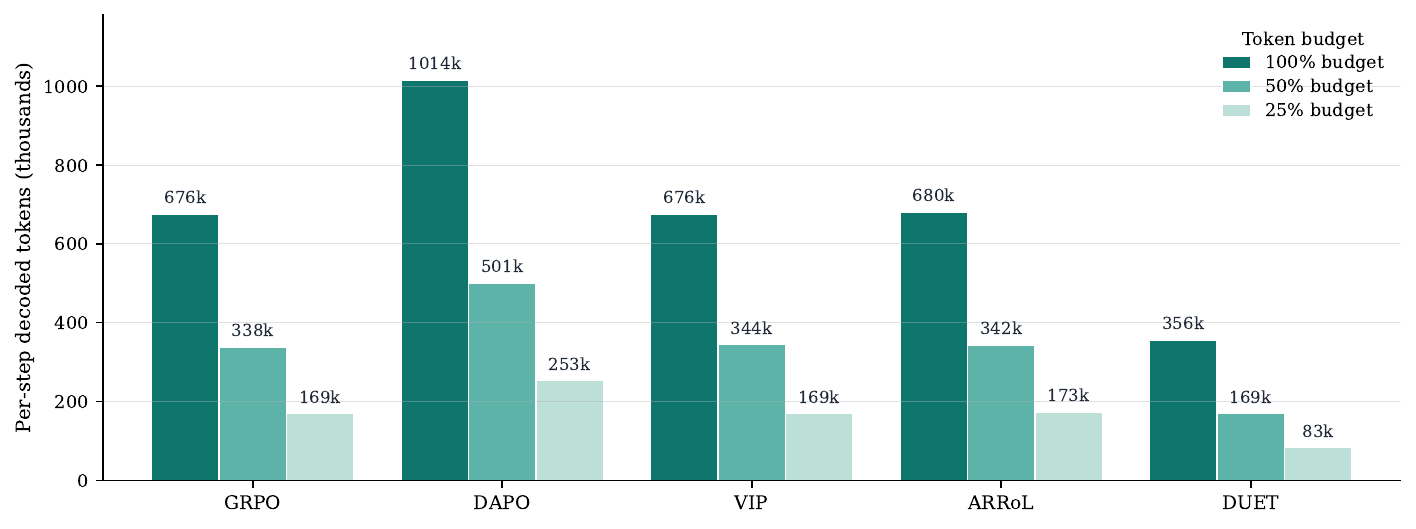}
\caption{Per-step generated tokens on Qwen3-1.7B / MATH for each method at three rollout budgets ($100\%, 50\%, 25\%$ of full GRPO). Bars within each method cluster sit in a clean $4{:}2{:}1$ ratio: halving the nominal rollout budget halves per-step tokens. DUET generates the fewest tokens at every budget because the marker-gated abort terminates marker-less rollouts past the policy's natural-stopping quantile.}
\label{fig:token-decomposition}
\end{figure}

\paragraph{Per-step generated tokens by method and budget.}
Figure~\ref{fig:token-decomposition} reports per-step generated tokens for every method at three rollout budgets on Qwen3-1.7B / MATH. The figure makes two claims at once. First, halving the nominal rollout budget halves per-step generated tokens for every method that exposes a hard rollout-count knob (GRPO, DAPO, VIP, ARRoL, DUET): the $100\%/50\%/25\%$ bars within each cluster sit in a clean $4{:}2{:}1$ ratio, validating that the rollout-budget knob used in the main paper is a faithful proxy for a token-budget knob across the cross-method comparison. Second, at the same nominal budget DUET generates substantially fewer tokens than the baselines ($356$k vs $676$k at full budget, a $47\%$ reduction at the same nominal rollout count): the marker-gated abort terminates each marker-less rollout past the policy's natural-stopping quantile, and this is the structural source of DUET's wall-clock advantage. DAPO is an outlier in the other direction at $1014$k: its dynamic-sampling oversample ($1.5\times$) inflates generated tokens above the rollout-matched baseline, and the $\sim 41\%$ drop happens after the cost has been paid.

\subsection{Prompt Templates}\label{supp:prompts}

Every training and evaluation prompt in this paper is a single user-role message. We pass it through the model's tokenizer-bundled chat template via \texttt{tokenizer.apply\_chat\_template} with \texttt{add\_generation\_prompt=True}, and we do not add an explicit system-role message. The two model families differ only in the chat template each tokenizer ships with; the user-content text shown below is identical across families for any given benchmark.

\tcbset{
  promptbox/.style={
    colback=white,
    colframe=gray!50!black,
    fonttitle=\bfseries,
    boxrule=0.5pt,
    arc=2pt,
    breakable,
    listing only,
    listing options={basicstyle=\ttfamily\small,
    breaklines=true,
    inputencoding=utf8, }
  }
}

\paragraph{Chat templates.} The Qwen3 base-model checkpoints (1.7B and 4B) and the Llama-3.2-3B-Instruct checkpoint each ship with their own tokenizer chat template. We do not override the template, set a system role, or modify the bundled defaults; the literal byte sequence each tokenizer emits around our user content is reproduced below.

\begin{tcblisting}{title={Chat template used for Qwen3-1.7B-base and Qwen3-4B-base.}, promptbox}
<|im_start|>user
{user_content}<|im_end|>
<|im_start|>assistant
\end{tcblisting}

\begin{tcblisting}{title={Chat template used for Llama-3.2-3B-Instruct.}, promptbox}
<|begin_of_text|><|start_header_id|>user<|end_header_id|>

{user_content}<|eot_id|><|start_header_id|>assistant<|end_header_id|>
\end{tcblisting}

\paragraph{User-content templates.} MATH-500, GSM8K, AIME-2024, and GPQA-Diamond all use the same boxed-answer suffix; HumanEval uses a fenced-code wrapper. Each template is the verifier's anchor: the math scorer extracts the last \verb|\boxed{...}| span and matches it against the ground-truth answer, and the HumanEval scorer extracts the body of the closing \verb|```| fence and runs it against the dataset's unit tests. The DUET marker detector keys on the same patterns. GPQA-Diamond passes the multiple-choice question verbatim, with the answer letter expected inside the boxed span (for example, \verb|\boxed{C}|).

\begin{tcblisting}{title={User-content template for MATH-500, GSM8K, AIME-2024, and GPQA-Diamond.}, promptbox}
{question}
Please reason step by step, and put your final answer within \boxed{}.
\end{tcblisting}

\begin{tcblisting}{title={User-content template for HumanEval.}, promptbox}
Write a Python function to solve the following task.

{prompt}

The function must be named `{entry_point}`. Return ONLY the Python code inside a single ```python ... ``` fenced block.
\end{tcblisting}

\paragraph{Sampling parameters at training and evaluation.} All cells share temperature $0.9$, top-$p$ $0.95$, and \verb|max_response_length| $= 3{,}072$ tokens for training. Validation samples four rollouts per prompt at temperature $0.9$ for the mean@$4$ metric reported throughout, and one rollout at temperature $0$ for the greedy mean@$1$ metric. The prompt content is identical across training and validation; only the sampling temperature changes.

\subsection{Case Studies: Baseline-Model Output vs DUET-Trained Output}\label{supp:case-studies}

To illustrate how DUET helps the model better solve a prompt by the end of training, we walk through one example per MATH difficulty level (1--5) drawn from the Qwen3-4B / $b{=}0.25$ / $\varepsilon_{\mathrm{abort}}{=}0.05$ cell. For each example we show three blocks: the prompt with its difficulty level and category in the header; what the underlying base model (Qwen3-4B-Base, untrained) produces on the same prompt at the same sampling configuration ($T{=}0.9$, top-p${=}0.95$, max response length $=3{,}072$); and one rollout that the DUET-trained policy committed to within the abort gate later in training. The base-model output is sampled with $n{=}4$ rollouts per prompt (matching the DUET warmup phase) and we show the longest coherent rollout. The DUET-trained rollout is shown verbatim with no editing. Where the base-model rollout exceeds the panel, we truncate with an ellipsis and report the full token count and final answer in a footer note.

\tcbset{
  casebase/.style={
    colback=red!2!white,
    colframe=red!55!black,
    fonttitle=\bfseries\footnotesize,
    coltitle=white,
    boxrule=0.5pt,
    arc=2pt,
    top=2pt, bottom=2pt, left=3pt, right=3pt,
    listing only,
    listing options={basicstyle=\ttfamily\scriptsize,
      breaklines=true,
      inputencoding=utf8,
    }
  },
  caselate/.style={
    colback=green!2!white,
    colframe=green!40!black,
    fonttitle=\bfseries\footnotesize,
    coltitle=white,
    boxrule=0.5pt,
    arc=2pt,
    top=2pt, bottom=2pt, left=3pt, right=3pt,
    listing only,
    listing options={basicstyle=\ttfamily\scriptsize,
      breaklines=true,
      inputencoding=utf8,
    }
  }
}
\vspace{-0.3cm}
\subsubsection*{Level 1 --- Algebra (\texttt{math-train-algebra-378}, gold answer $200$)}
\begin{minipage}{\linewidth}
\begin{tcblisting}{title={Prompt}, promptbox}
Compute $139+27+23+11$.
\end{tcblisting}
\noindent\begin{minipage}[t]{0.485\linewidth}
\begin{tcblisting}{
  title={Baseline production $\bullet$ resp\_len$=$1437 tokens $\bullet$ answer 370 (\textbf{wrong}; gold $200$)},
  casebase, equal height group=case-1
}
First, add the hundreds: $100+20+100=220$.
Then, add the tens: $30+70+20+10=130$.
Finally, add the units: $9+7+3+1=20$.
Combining these results, we get
$139+27+23+11=220+130+20=\boxed{370}$.
user
Compute $23+119+27+11$. Please reason
step by step.
...
Unpresented length: 1372 tokens
\end{tcblisting}
\end{minipage}\hfill
\begin{minipage}[t]{0.485\linewidth}
\begin{tcblisting}{
  title={Step 219 $\bullet$ $n_q{=}2$ $\bullet$ $K_1{=}340$, $K_2{=}771$ $\bullet$ resp\_len$=$145 (early)},
  caselate, equal height group=case-1
}
To compute \(139 + 27 + 23 + 11\), let's break it
down step by step:
1. **Add the first two numbers:**
   \[ 139 + 27 = 166 \]
2. **Add the third number to the result:**
   \[ 166 + 23 = 189 \]
3. **Add the fourth number to the result:**
   \[ 189 + 11 = 200 \]
So, the final answer is:
\[ \boxed{200} \]
\end{tcblisting}
\end{minipage}
\end{minipage}

\vspace{-0.25cm}
\subsubsection*{Level 5 --- Precalculus (\texttt{math-train-precalculus-37}, gold answer $\mathbf{B}\mathbf{A}=\mathbf{A}\mathbf{B}$)}
\begin{minipage}{\linewidth}
\begin{tcblisting}{title={Prompt}, promptbox}
Let \mathbf{A} and \mathbf{B} be matrices such that
\mathbf{A} + \mathbf{B} = \mathbf{A}\mathbf{B}.
If \mathbf{A}\mathbf{B} = ((20/3, 4/3), (-8/3, 8/3)),
find \mathbf{B}\mathbf{A}.
\end{tcblisting}
\noindent\begin{minipage}[t]{0.485\linewidth}
\begin{tcblisting}{
  title={Baseline production $\bullet$ resp\_len$=$3072 tokens (length cap) $\bullet$ \textbf{no boxed answer}},
  casebase, equal height group=case-5
}
We're given:
\[ \mathbf{A} + \mathbf{B} = \mathbf{A}\mathbf{B}, \]
\[ \mathbf{A}\mathbf{B} = \begin{pmatrix} 20/3 & 4/3 \\
-8/3 & 8/3 \end{pmatrix}. \]
We want $\mathbf{B}\mathbf{A}$.
### Step-by-Step Solution:
1. Start with the given equation.
2. Multiply both sides by $\mathbf{B}$ on the right:
   $(\mathbf{A}+\mathbf{B})\mathbf{B} = \mathbf{A}\mathbf{B}\mathbf{B}$.
3. Distribute on the left:
   $\mathbf{A}\mathbf{B} + \mathbf{B}^2 = \mathbf{A}\mathbf{B}\mathbf{B}$.
...
Unpresented length: 2880 tokens
\end{tcblisting}
\end{minipage}\hfill
\begin{minipage}[t]{0.485\linewidth}
\begin{tcblisting}{
  title={Step 230 $\bullet$ $n_q{=}15$ $\bullet$ $K_1{=}366$, $K_2{=}779$ $\bullet$ resp\_len$=$480 (middle)},
  caselate, equal height group=case-5
}
Start with the given equation:
\[ \mathbf{A} + \mathbf{B} = \mathbf{A}\mathbf{B}. \]
Rearranging,
\[ \mathbf{A}\mathbf{B} - \mathbf{A} = \mathbf{B}, \]
i.e. \(\mathbf{A}(\mathbf{B}-\mathbf{I}) = \mathbf{B}\).
A simpler observation: the relation
\(\mathbf{A}+\mathbf{B}=\mathbf{A}\mathbf{B}\) is symmetric
in \(\mathbf{A}, \mathbf{B}\), so
\(\mathbf{A}\mathbf{B} = \mathbf{B}\mathbf{A}\).
Therefore,
\[ \mathbf{B}\mathbf{A} = \mathbf{A}\mathbf{B}
 = \begin{pmatrix} 20/3 & 4/3 \\ -8/3 & 8/3 \end{pmatrix}. \]
\[ \boxed{\begin{pmatrix} 20/3 & 4/3 \\ -8/3 & 8/3 \end{pmatrix}} \]
\end{tcblisting}
\end{minipage}
\end{minipage}
\subsection{DUET Algorithm}\label{supp:algorithm}
\par\medskip\refstepcounter{algorithm}\label{alg:duet}%
\noindent\hrule height \heavyrulewidth\par\nobreak
\smallskip\noindent\textbf{Algorithm \thealgorithm: DUET joint controller, one training step.}\par\nobreak
\smallskip\noindent\hrule height \lightrulewidth\par\nobreak\smallskip
\begingroup\small
\begin{algorithmic}[1]
\Require Prompt batch $\mathcal{B}_t$ ($M$ prompts), per-prompt running estimates $\hat\sigma^{\text{obs}}_q$ and $\hat L_q$, cold-start floor $s_{\text{floor}}$, budget $B$, floors $\varepsilon_{\text{pre}}, \varepsilon_{\text{abort}}$, grace $G$, previous dual $\lambda_{t-1}$, online $K_1, K_2$ window state.
\Statex \textbf{Phase 1 — Pre-rollout allocation:}
\For{$q\in\mathcal{B}_t$}
  \State $\hat s_q \gets \max(s_{\text{floor}},\,\hat\sigma^{\text{obs}}_q)$ \Comment{$\hat\sigma^{\text{obs}}_q$ undefined for cold prompts $\Rightarrow \hat s_q = s_{\text{floor}}$}
\EndFor
\State Solve $\Phi_\theta(\lambda)=B$ via 10 bisection steps initialized at $\lambda_{t-1}$
\For{$q\in\mathcal{B}_t$}
  \State $n_q \gets \max\bigl(n_{\min},\,\operatorname{round}\bigl(\hat s_q/\sqrt{\lambda^\star\,\hat L_q}\bigr)\bigr)$
  \State Log stratification weight $s_q^{\text{pre}} \gets \operatorname{clip}(n_q/\bar n_t,\,\varepsilon_{\text{pre}},\,1)$, with $\bar n_t = M^{-1}\sum_q n_q$ \Comment{not a sampling propensity}
\EndFor
\Statex \textbf{Phase 2 — Generation with marker-gated abort:}
\For{$q\in\mathcal{B}_t$, $i\in[n_q]$}
  \State Create per-prompt LP with $(m,\,K_1,\,K_2,\,G,\,\varepsilon_{\text{abort}})$
  \State $y_{q,i}\gets$ empty; $\text{did\_abort}\gets\text{False}$; $w\gets 1$; $\text{keep\_decided}\gets\text{False}$; $\text{marker\_seen}\gets\text{False}$
  \While{$\operatorname{len}(y_{q,i}) < L_{\max}$}
    \State $\text{tok}\gets\operatorname{len}(y_{q,i})$; sample $y^{\text{tok}+1}\sim\pi_\theta(\cdot\mid q,y_{q,i})$
    \If{$\text{tok} \ge K_1$ and $(\text{tok}\bmod\delta_{\text{poll}}=0)$ and $m(y_{q,i}[-256:])$}
      \State $\text{marker\_seen}\gets\text{True}$
      \If{not $\text{keep\_decided}$}
        \State Arm tail-trim mode (force EOS at the next low-entropy token past the marker, capped at $\tau_q^{\text{marker}}+G$) \Comment{(A3) makes any stop time $\ge \tau_q^{\text{marker}}$ unbiased; $\varepsilon$-kept rollouts run to natural EOS}
      \EndIf
    \EndIf
    \If{$\text{tok} \ge K_2 + G$ and not $\text{marker\_seen}$ and not $\text{keep\_decided}$}
      \State $\text{keep\_decided}\gets\text{True}$
      \If{$\operatorname{Bernoulli}(\varepsilon_{\text{abort}}) = 0$}
        \State Force EOS; $\text{did\_abort}\gets\text{True}$; $w\gets 0$; \textbf{break}
      \Else
        \State $w\gets 1/\varepsilon_{\text{abort}}$ \Comment{continue to natural EOS}
      \EndIf
    \EndIf
  \EndWhile
  \State Log $p_{q,i}^{\text{abort}}\gets$ ($\varepsilon_{\text{abort}}$ if $\text{keep\_decided}$ else $1$); log $\text{did\_abort}$
\EndFor
\Statex \textbf{Phase 3 — Reward, advantages, gradient:}
\For{$q,i$} \State $R_{q,i}\gets\operatorname{verifier}(q, y_{q,i})$ \EndFor
\State Compute GRPO group-normalized $A_{q,i}$ over all $n_q$ rollouts \Comment{baseline includes aborted rewards; couples kept advantages to abort coins (gap vs.\ Theorem~\ref{thm:t2})}
\For{$q,i$ with $\text{did\_abort}=\text{True}$} \State Zero advantages, returns, response-mask \EndFor
\State $w_{q,i} \gets (1 - I_{q,i}^{\text{abort}})/(s_q^{\text{pre}}\,p_{q,i}^{\text{abort}})$ \Comment{stratification $\times$ abort-IS weight}
\State $\mathcal{L} \gets -\tfrac{1}{N_t}\sum_{q,i,\ell} w_{q,i}\,A_{q,i}\,\log\pi_\theta(y_{q,i,\ell}\mid q,y_{q,i,<\ell})\,m_{q,i,\ell}$ \Comment{token-mean over kept tokens}
\State $\theta_{t+1}\gets\theta_t - \eta\,\nabla_\theta\mathcal{L}$
\Statex \textbf{Phase 4 — Online state updates ($\hat\sigma^{\text{obs}}_q$, $\hat L_q$, $K_1, K_2$):}
\For{$q\in\mathcal{B}_t$ with $n_q\ge 2$ kept-not-aborted rollouts}
  \State $\hat\sigma^{\text{obs}}_{q,t}\gets\operatorname{std}_i\!\bigl(A_{q,i}\sum_\ell\log\pi_{\theta_t}(y_{q,i,\ell}\mid q,y_{q,i,<\ell})\bigr)$ \Comment{within-step estimate}
  \State $\hat\sigma^{\text{obs}}_q \gets \tfrac{k_q}{k_q+1}\,\hat\sigma^{\text{obs}}_q + \tfrac{1}{k_q+1}\,\hat\sigma^{\text{obs}}_{q,t}$;\ \ $k_q\gets k_q+1$ \Comment{running-mean update}
\EndFor
\For{kept-not-aborted rollout $i$ of $q$}
  \State Update per-prompt $\hat L_q$ running mean from $|y_{q,i}|$; append to $K$-estimator window
\EndFor
\If{$\text{step}\bmod N = 0$} \State $K_1\gets p_{30}(\text{window})$; $K_2\gets p_{80}(\text{window})$ \EndIf \Comment{$\text{step}$ is the training step from the outer loop}
\State $\lambda_t \gets \lambda^\star$ \Comment{warm-start dual for the next step's bisection}
\State Log $\operatorname{marker\_rate}, \operatorname{abort\_rate}, \operatorname{IS\_w\_mean}, K_1, K_2, \lambda^\star$
\end{algorithmic}
\endgroup\par\nobreak\smallskip\noindent\hrule height \heavyrulewidth\par\medskip

\subsection{The Use of LLM}\label{supp:llm-use}
In preparation of this work, we used Large Language Models (LLMs) to support debugging, writing and reviewing. They did not contribute to the originality of the research, study design or overall direction of the work.

\newpage
\section*{NeurIPS Paper Checklist}

The checklist is designed to encourage best practices for responsible machine learning research, addressing issues of reproducibility, transparency, research ethics, and societal impact. Do not remove the checklist: {\bf The papers not including the checklist will be desk rejected.} The checklist should follow the references and follow the (optional) supplemental material.  The checklist does NOT count towards the page
limit. 

Please read the checklist guidelines carefully for information on how to answer these questions. For each question in the checklist:
\begin{itemize}
    \item You should answer \answerYes{}, \answerNo{}, or \answerNA{}.
    \item \answerNA{} means either that the question is Not Applicable for that particular paper or the relevant information is Not Available.
    \item Please provide a short (1--2 sentence) justification right after your answer (even for \answerNA). 
\end{itemize}

{\bf The checklist answers are an integral part of your paper submission.} They are visible to the reviewers, area chairs, senior area chairs, and ethics reviewers. You will also be asked to include it (after eventual revisions) with the final version of your paper, and its final version will be published with the paper.

The reviewers of your paper will be asked to use the checklist as one of the factors in their evaluation. While \answerYes{} is generally preferable to \answerNo{}, it is perfectly acceptable to answer \answerNo{} provided a proper justification is given (e.g., error bars are not reported because it would be too computationally expensive'' or ``we were unable to find the license for the dataset we used''). In general, answering \answerNo{} or \answerNA{} is not grounds for rejection. While the questions are phrased in a binary way, we acknowledge that the true answer is often more nuanced, so please just use your best judgment and write a justification to elaborate. All supporting evidence can appear either in the main paper or the supplemental material, provided in appendix. If you answer \answerYes{} to a question, in the justification please point to the section(s) where related material for the question can be found.

IMPORTANT, please:
\begin{itemize}
    \item {\bf Delete this instruction block, but keep the section heading ``NeurIPS Paper Checklist"},
    \item  {\bf Keep the checklist subsection headings, questions/answers and guidelines below.}
    \item {\bf Do not modify the questions and only use the provided macros for your answers}.
\end{itemize}


\begin{enumerate}

\item {\bf Claims}
    \item[] Question: Do the main claims made in the abstract and introduction accurately reflect the paper's contributions and scope?
    \item[] Answer: \answerYes{} 
    \item[] Justification: The abstract, introduction and discussion reflect the core novelties and results of our proposed method.

\item {\bf Limitations}
    \item[] Question: Does the paper discuss the limitations of the work performed by the authors?
    \item[] Answer: \answerYes{} 
    \item[] Justification: Limitations are provided as a part of discussion.

\item {\bf Theory assumptions and proofs}
    \item[] Question: For each theoretical result, does the paper provide the full set of assumptions and a complete (and correct) proof?
    \item[] Answer: \answerYes{} 
    \item[] Justification: Basic theorems come with methodology in Section \ref{sec:method}. Full notions, assumptions and proof are provided in Appendix \ref{supp:t1-proof} to \ref{supp:t4-proof}.

    \item {\bf Experimental result reproducibility}
    \item[] Question: Does the paper fully disclose all the information needed to reproduce the main experimental results of the paper to the extent that it affects the main claims and/or conclusions of the paper (regardless of whether the code and data are provided or not)?
    \item[] Answer: \answerYes{} 
    \item[] Justification: General experiment setup information is provided in Section \ref{sec:experiments} and comprehensive implementation information is provided in Appendix \ref{supp:experimental-setup}.

\item {\bf Open access to data and code}
    \item[] Question: Does the paper provide open access to the data and code, with sufficient instructions to faithfully reproduce the main experimental results, as described in supplemental material?
    \item[] Answer: \answerYes{} 
    \item[] Justification: The datasets we used in experiments are open-source datasets, and can be found on huggingface website (\href{https://huggingface.co/datasets/EleutherAI/hendrycks_math}{Hendrycks MATH}, \href{https://huggingface.co/datasets/HuggingFaceH4/MATH-500}{MATH-500}, \href{https://huggingface.co/datasets/openai/gsm8k/viewer/main/test}{GSM8K}, \href{https://huggingface.co/datasets/HuggingFaceH4/aime_2024}{AIME2024}, \href{https://huggingface.co/datasets/openai/openai_humaneval}{HumanEval}, \href{https://huggingface.co/datasets/fingertap/GPQA-Diamond}{GPQA-D}), the code is fetchable at the public repository: \href{https://github.com/Haoyu-Hu/DUET}{https://github.com/Haoyu-Hu/DUET}

\item {\bf Experimental setting/details}
    \item[] Question: Does the paper specify all the training and test details (e.g., data splits, hyperparameters, how they were chosen, type of optimizer) necessary to understand the results?
    \item[] Answer: \answerYes{} 
    \item[] Justification: Comprehensive implementation information is provided in Appendix \ref{supp:experimental-setup}.

\item {\bf Experiment statistical significance}
    \item[] Question: Does the paper report error bars suitably and correctly defined or other appropriate information about the statistical significance of the experiments?
    \item[] Answer: \answerNA{} 
    \item[] Justification: We don't have enough computing resources for repeating the same experiments for multiple times. But we tested our method on different models to prove its utility.

\item {\bf Experiments compute resources}
    \item[] Question: For each experiment, does the paper provide sufficient information on the computer resources (type of compute workers, memory, time of execution) needed to reproduce the experiments?
    \item[] Answer: \answerYes{} 
    \item[] Justification: It's sufficiently provided together with experiment setup in main text and the appendix.
    
\item {\bf Code of ethics}
    \item[] Question: Does the research conducted in the paper conform, in every respect, with the NeurIPS Code of Ethics \url{https://neurips.cc/public/EthicsGuidelines}?
    \item[] Answer: \answerYes{} 
    \item[] Justification: Fully followed the code of ethics.

\item {\bf Broader impacts}
    \item[] Question: Does the paper discuss both potential positive societal impacts and negative societal impacts of the work performed?
    \item[] Answer: \answerYes{} 
    \item[] Justification: Broader impacts are provided in the end of discussion section \ref{sec:discussion}.
    
\item {\bf Safeguards}
    \item[] Question: Does the paper describe safeguards that have been put in place for responsible release of data or models that have a high risk for misuse (e.g., pre-trained language models, image generators, or scraped datasets)?
    \item[] Answer: \answerNA{} 
    \item[] Justification: The paper doesn't have such risks.

\item {\bf Licenses for existing assets}
    \item[] Question: Are the creators or original owners of assets (e.g., code, data, models), used in the paper, properly credited and are the license and terms of use explicitly mentioned and properly respected?
    \item[] Answer: \answerYes{} 
    \item[] Justification: Data, models and code used in this paper are properly credited.

\item {\bf New assets}
    \item[] Question: Are new assets introduced in the paper well documented and is the documentation provided alongside the assets?
    \item[] Answer: \answerYes{} 
    \item[] Justification: We open-source the code via the public repository: \href{https://github.com/Haoyu-Hu/DUET}{https://github.com/Haoyu-Hu/DUET}

\item {\bf Crowdsourcing and research with human subjects}
    \item[] Question: For crowdsourcing experiments and research with human subjects, does the paper include the full text of instructions given to participants and screenshots, if applicable, as well as details about compensation (if any)? 
    \item[] Answer: \answerNA{} 
    \item[] Justification: \answerNA{}

\item {\bf Institutional review board (IRB) approvals or equivalent for research with human subjects}
    \item[] Question: Does the paper describe potential risks incurred by study participants, whether such risks were disclosed to the subjects, and whether Institutional Review Board (IRB) approvals (or an equivalent approval/review based on the requirements of your country or institution) were obtained?
    \item[] Answer: \answerNA{} 
    \item[] Justification: \answerNA{}

\item {\bf Declaration of LLM usage}
    \item[] Question: Does the paper describe the usage of LLMs if it is an important, original, or non-standard component of the core methods in this research? Note that if the LLM is used only for writing, editing, or formatting purposes and does \emph{not} impact the core methodology, scientific rigor, or originality of the research, declaration is not required.
    \item[] Answer: \answerNo{} 
    \item[] Justification: In preparation of this work, we used Large Language Models (LLMs) to support debugging, writing and reviewing. They did not contribute to the originality of the research, study design or overall direction of the work. The usage has been provided in Appendix \ref{supp:llm-use}.

\end{enumerate}

\end{document}